\documentclass[10pt,twocolumn,letterpaper]{article}

\usepackage{3dv}
\usepackage{times}
\usepackage{epsfig}
\usepackage{graphicx}
\usepackage{amsmath}
\usepackage{amssymb}
\usepackage{amsthm}
\usepackage{xcolor}
\usepackage{multirow}
\usepackage{verbatim} 
\usepackage{authblk}
\usepackage[ruled,linesnumbered]{algorithm2e}
\DeclareFontFamily{U}{BOONDOX-calo}{\skewchar\font=45 }
\DeclareFontShape{U}{BOONDOX-calo}{m}{n}{
  <-> s*[1.05] BOONDOX-r-calo}{}
\DeclareFontShape{U}{BOONDOX-calo}{b}{n}{
  <-> s*[1.05] BOONDOX-b-calo}{}
\DeclareMathAlphabet{\mathcalboondox}{U}{BOONDOX-calo}{m}{n}
\SetMathAlphabet{\mathcalboondox}{bold}{U}{BOONDOX-calo}{b}{n}
\DeclareMathAlphabet{\mathbcalboondox}{U}{BOONDOX-calo}{b}{n}

\usepackage{subcaption}
\DeclareMathOperator*{\argmax}{arg\,max}
\usepackage[pagebackref=true,breaklinks=true,colorlinks,bookmarks=false]{hyperref}

\threedvfinalcopy % *** Uncomment this line for the final submission

\def\threedvPaperID{..} % *** Enter the 3DV Paper ID here

\definecolor{darkgreen}{rgb}{0,0.5,0} % define darkgreen color
\definecolor{purple}{rgb}{1,0,1} % define purple color

\newcommand{\cadops}{\emph{\hbox{CADOps-Net}}}
\newcommand{\ccdops}{\emph{\hbox{CC3D-Ops}}}
\newcommand{\opt}{\textit{\hbox{op.type}}}
\newcommand{\opts}{\textit{\hbox{op.types}}}
\newcommand{\ops}{\textit{\hbox{op.step}}}
\newcommand{\opss}{\textit{\hbox{op.steps}}}

\begin{document}

%%%%%%%%% TITLE
\title{CADOps-Net: Jointly Learning CAD Operation Types and Steps from Boundary-Representations
}
\setlength{\affilsep}{0.1em}
\renewcommand\Affilfont{\tt\small}
\author[1]{Elona Dupont}
\author[1, 2]{Kseniya Cherenkova}
\author[1]{Anis Kacem}
\author[1]{Sk Aziz Ali}
\author[2]{Ilya Arzhannikov}
\author[2]{Gleb Gusev}
\author[1]{Djamila Aouada}
\affil[1]{SnT, University of Luxembourg, Luxembourg}
\affil[2]{Artec 3D, Luxembourg}

%
% \thispagestyle{empty}

%%%%%%%%% ABSTRACT
\maketitle

\begin{abstract}
3D reverse engineering is a long sought-after, yet not completely achieved goal in the Computer-Aided Design (CAD) industry. The objective is to recover the construction history of a CAD  model.
Starting from a \textit{Boundary Representation (B-Rep)} of a CAD model, this paper proposes a new deep neural network, \textbf{\textit{CADOps-Net}}, that jointly learns the CAD operation types and the decomposition into different CAD operation steps. This joint learning allows to divide a B-Rep into parts that were created by various types of CAD operations at the same construction step; therefore providing relevant information for further recovery of the design history. 
Furthermore, we propose the novel \textbf{\textit{CC3D-Ops}} dataset that includes over $37k$ CAD models annotated with CAD operation type labels and step labels. Compared to existing datasets, the complexity and variety of CC3D-Ops models are closer to those used for industrial purposes. Our experiments, conducted on the proposed CC3D-Ops and the publicly available Fusion360 datasets, demonstrate the competitive performance of CADOps-Net with respect to state-of-the-art, and confirm the importance of the joint learning of CAD operation types and steps.
\end{abstract}

%%%%%%%%%%%%%%%%%%%%%%%%%%%%%%%%%%%%%%%%%%%%%%%%%%%%%%
\vspace{-0.2cm}
\section{Introduction}

\begin{figure}[ht]
\begin{center}
\includegraphics[width=0.9\linewidth]{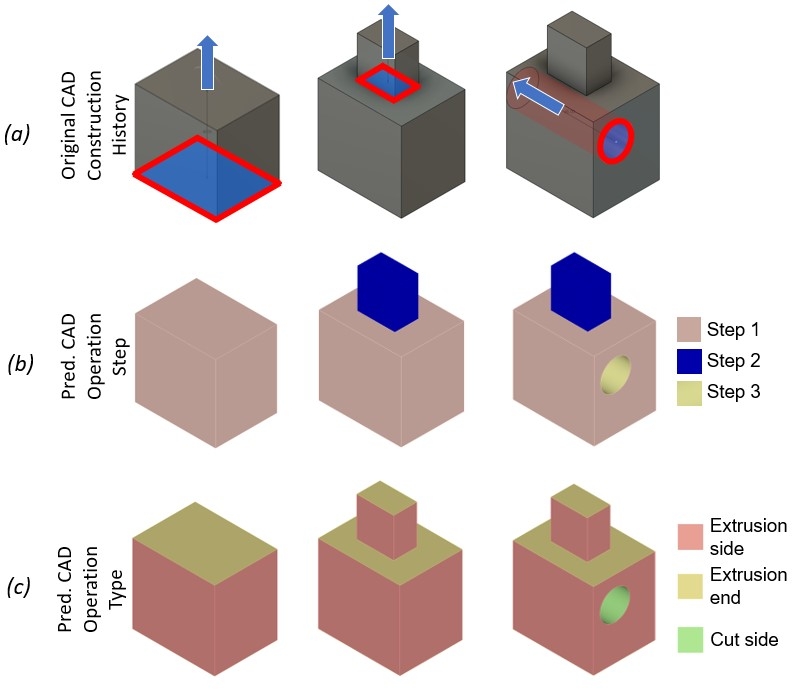}
\end{center}
 \caption{B-Rep segmentation into CAD operations types and steps.}
\label{fig:seg-group}
\end{figure}
In today’s digital era, Computer-Aided Design (CAD) is the standard option for designing objects ahead of manufacturing~\cite{drawing,importance,cadhistory}.
The parametric nature of CAD models allows engineers and designers to iterate over the parameters of existing CAD models to edit and adapt them to new contexts, such as customizing dental prostheses~\cite{solaberrieta2014computer}, or modifying mechanical parts~\cite{thompson1999feature}. However, this is only possible if the final shape of the CAD model comes with its design history. Unfortunately, this is rarely the case as the design history is often not available for generic 3D shapes~\cite{inversecsg} or lost when CAD models are exchanged between different CAD applications~\cite{kim2007integration,brepnet}. Consequently, the research community has put a lot of efforts in relating the geometry of 3D shapes to the CAD design history~\cite{brepnet,uvnet,csgnet,inversecsg,zonegraph,deepcad}. This process is known as \textit{3D~reverse~engineering}. 

Prior works attempted to recover the CAD design history, considering \textit{Constructive Solid Geometry} (CSG) based models~\cite{inversecsg,csgnet} for simplicity. In CSG, a CAD model is represented by a set of rigidly transformed solid primitives (\eg cube, sphere, cylinder) and combined using Boolean operations such as union, intersection, and difference~\cite{csg-brep}. 
However, modern CAD workflows use \textit{feature-based} modeling, in which solids are created by iteratively adding features such as holes, slots, or bosses~\cite{zonegraph,FeatureNet}. These high-level features are sequentially created through drawing~\textit{sketches} and applying \textit{CAD operations} such as `\textit{extrusion}', `\textit{revolution}', etc. Figure~\ref{fig:seg-group}\textcolor{red}{a} illustrates an example of feature-based simple CAD model creation. Using this type of CAD modeling, the final model is stored in a data structure called \textit{Boundary-Representation} (B-Rep). The B-Rep describes the geometry and the topology of the CAD model through faces, edges, loops, co-edges and vertices~\cite{brepnet}. However, it does not include information about how these entities are designed.
Accordingly, recent efforts in the state-of-the-art have focused on relating B-Reps to the design history~\cite{zonegraph,brepnet,uvnet}. In particular, two main directions have been followed: (1)~segmenting the B-Rep faces into CAD operation types (\eg `\textit{extrusion}', `\textit{revolution}')~\cite{uvnet,brepnet} or higher-level machining features (\eg `\textit{holes}', `\textit{slots}')~\cite{cadnet} that allowed their creation; (2)~inferring a sequence of parametric sketches and extrusions that allowed the design of the B-Rep~\cite{zonegraph,fusion360,point2cyl}. While the first group of works have the advantage of relating each face of the B-Rep to various types of CAD operations, they do not describe the relationship between the faces nor the steps of the construction. On the other hand, the works taking the second direction reconstruct the ordered sequence of the design history, including sketches, but they are usually limited to only one CAD operation type (\ie `\textit{extrusion}') as a simplification of the search space. 

In this work, we combine both directions by segmenting the faces of the B-Reps into various CAD operation types and further decomposing them into steps of construction as shown in~Figure~\ref{fig:seg-group}. These two aspects are jointly learned using an end-to-end neural network, allowing the recovery of further information about the design history such as CAD sketches. The proposed method is evaluated on the publicly available Fusion360 dataset~\cite{fusion360}, and a newly introduced dataset that is closer to real-world challenges. The key \textbf{contributions} can be summarized as follows:
\vspace{-0.15cm}
\begin{itemize}
  \item A neural network, \cadops, that operates on B-Reps is proposed to learn the segmentation of faces into CAD operation types and steps. We introduce a joint learning method within an end-to-end model.
  \vspace{-0.15cm}
  \item We create a novel dataset, \ccdops, that builds on top of the existing CC3D~dataset~\cite{cc3d} by extending it with B-Reps and their corresponding per-face CAD operation type and step annotations. Compared to existing datasets~\cite{fusion360,mfcad,abc_dataset}, \ccdops~better reflects real-world industrial challenges thanks to the complexity of its CAD models. This dataset can be found at \url{https://cvi2.uni.lu/cc3d-ops/}. 
  \vspace{-0.15cm}
  \item The proposed approach is evaluated on two datasets and compared to recent state-of-the-art methods. We further showcase some preliminary results on a possible downstream application consisting of CAD sketch recovery from B-Reps.
  
\end{itemize}

The rest of the paper is organized as follows; In Section~\ref{sec:related-works} related works are discussed followed by the problem formulation in  Section~\ref{sect:pblmstatement}. Section~\ref{sec:approach} describes the proposed \cadops. The proposed \ccdops~dataset is introduced in Section~\ref{dataset}. The experimental results are reported and analyzed in Section~\ref{sec:expt}. Finally, Section~\ref{sec:conclusion} concludes this work and presents directions for future work. 

%%%%%%%%%%%%%%%%%%%%%%%%%%%%%%%%%%%%%%%%%%%%%%%%%%%%%%
\section{Related Works}
\label{sec:related-works}
Learning representations for 3D shape modeling~\cite{ahmed2018survey} is an important research topic that aims at finding the best deep feature encoding method. For instance, while a group of works leverages feature embedding for unordered and irregular point clouds~\cite{pointnet, pointnet++, dgcnn, PointConv, DensePoint} or regular grids of voxels~\cite{voxnet, pvcnn, cc3d, RPSRNet, o-cnn}, another group of works~\cite{meshCNN, spiralnet++, GeodscCNN} defines convolution kernels and feature embedding techniques for meshes and manifolds. Other works~\cite{inversecsg,brepnet,uvnet,csgnet} focused on learning from high-level 3D shape representations such as CAD models. These methods either assume that the CAD models are obtained using CSG or feature-based modeling. In particular, the recovery of the CAD design history considering these two types of modeling has attracted a lot of attention~\cite{fusion360,uvnet,brepnet, csgnet, inversecsg}.

\vspace{0.12cm}
\noindent\textbf{CSG-based Approaches.} Several approaches~\cite{csgnet, csg-solid, csg-brep, constructingCSG} attempt to infer the design history of CAD models using CSG representation. For instance, when the input shape is a 3D point cloud, \cite{inversecsg} and~\cite{constructingCSG} convert it to the CSG tree (mainly binary-tree) of solid bodies which is a volumetric representation of simple geometrical primitives. Similarly, when the input is a B-Rep or a solid body,~\cite{BRep2CSG} and~\cite{csg-solid} describe unique CSG conversion steps (or vice-versa in~\cite{ csg-brep}).
The conversion reveals hierarchical steps involved in modeling solid bodies, whereas CAD models appear more as connected surface patches than volumetric solids~\cite{netgen}. Therefore, predicting CSG construction history may not reveal the actual CAD construction steps used in modern CAD workflows~\cite{zonegraph}. The latter mostly consider B-Reps instead of CSG and rely on feature-based modeling, which is addressed in our work. 

\vspace{0.12cm}
\noindent\textbf{Feature-based Approaches. }The methods that either directly learn the B-Rep structure of a CAD model~\cite{uvnet, brepnet, solidgen, zonegraph, cadnet} or predict sketches and CAD operations~\cite{deepcad, sketchgen, sketchgraphs, cad-as-a-lang}, are closely related to our work. The works in~\cite{sketchgen, cad-as-a-lang} propose generative models for CAD sketches with a focus on the constraints of sketch entities. Therefore, they do not consider the connection between constrained CAD sketches and operations. On the other hand, methods like SolidGen~\cite{solidgen}, BRepNet~\cite{brepnet}, UV-Net~\cite{uvnet}, CADNet~\cite{cadnet} put more emphasis on how to use the B-Rep data structure to obtain face embeddings followed by face segmentation, but obscuring the relation between the segmented faces and design steps. DeepCAD~\cite{deepcad}, Fusion360~\cite{fusion360} and Zone-graph~\cite{zonegraph} are the first set of methods, to the best of our knowledge, that relate parametric sketches and CAD operations proposing a generative model for CAD design. However, their models were restricted to only one type of CAD operations, namely extrusion. Finally, Point2Cyl~\cite{point2cyl} operates on point clouds to detect 2D sketches but is also limited to the CAD extrusion operation.  

\vspace{0.12cm}
\noindent\textbf{CAD Modeling Datasets. } Besides Fusion360~\cite{fusion360}, there are no datasets that provide both B-Reps and fully explicit construction history in standard format. For example, the ABC dataset~\cite{abc_dataset} provides $1M+$ CAD models with sparse construction history provided in Onshape proprietary format~\cite{fusion360}. On the other hand, the SketchGraphs dataset~\cite{sketchgraphs} contains a large number of sketch construction sequences but not the B-Reps. Both MFCAD~\cite{mfcad} and MFCAD++~\cite{cadnet} datasets contain B-Reps and machining feature labels. However, the samples are synthetic models and too simple to consider for industrial modeling tasks. CC3D dataset~\cite{cc3d} offers $50k+$ pairs of industrial CAD models as triangular meshes and their corresponding 3D scans, but without construction steps and B-Reps. \ccdops~supplements the CC3D dataset with these elements.
%----------------------------------------

\begin{figure*}[t]
\begin{center}
\includegraphics[width=0.87\linewidth]{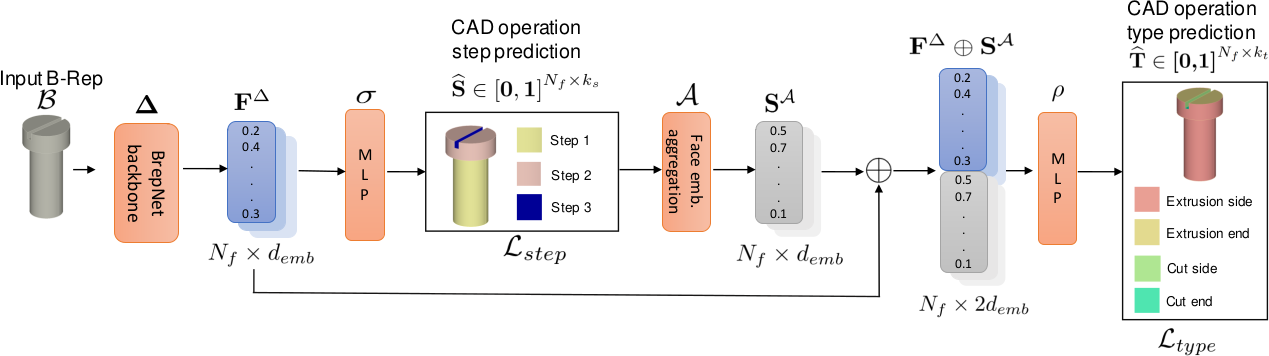}
   \caption{The \cadops~joint learning network architecture. The input B-Rep, $\mathcal{B}$, is first passed through a BrepNet backbone, $\mathbf{\Delta}$, to obtain face embeddings, $\mathbf{F}^{\Delta}$. These embeddings are then fed to an MLP layer, $\mathbf{\sigma}$, to predict the face \ops~segmentation, $\mathbf{\widehat{S}}$. Using these predictions, the face embeddings, $\mathbf{F}^{\Delta}$, are aggregated with a function $\mathcal{A}$ into step embeddings, $\mathbf{S}^{\mathcal{A}}$. Finally the concatenation, $\oplus$, of the face embeddings, $\mathbf{F}^{\Delta}$, and their corresponding step embeddings, $\mathbf{S}^{\mathcal{A}}$, are passed through an MLP layer, $\mathbf{\rho}$ to predict the \opt~face labels.}
\label{fig:pipeline}
\end{center}
\end{figure*}
%%%%%%%%%%%%%%%%%%%%%%%%%%%%%%%%%%%%%%%%%%%%%%%%%%%%%%

\section{Problem Statement}
\label{sect:pblmstatement}
A B-Rep $\mathcal{B}$ can be defined as a tuple of three sets of entities -- \ie, 
a set of $N_f$ faces \hbox{$\{f_1, f_2, \dots, f_{N_f}\}$},   
a set of $N_e$ edges \hbox{$\{e_1, e_2, \dots, e_{N_e}\}$}, 
and 
a set of $N_c$ co-edges (also known as directed half-edges) \hbox{$\{c_1, c_2, \dots, c_{N_c}\}$}. 
Our main goal is to relate each face $f$ in $\mathcal{B}$ with its construction history using three different types of features
\hbox{$\mathbf{F}\in\mathbb{R}^{N_f \times d_f}$}, \hbox{$\mathbf{E}\in\mathbb{R}^{N_e \times d_e}$},
and 
\hbox{$\mathbf{C}\in\mathbb{R}^{N_c \times d_c}$}
extracted for the three entities, namely, faces, edges, and co-edges, respectively\footnote{The considered features are described in Section~\ref{sect:exp-setup}.}.  
The CAD construction history is defined as a sequential combination of sketches followed by some CAD operations. In this work, we are interested in learning (1) the type of CAD operations through
the segmentation of each face that allowed for its creation, and (2) the CAD operation step to which the segmented face belongs.

\subsection{CAD Operation Types}
The choice of CAD operation types is crucial for constructing CAD models. For notation simplicity, let us denote them as \textbf{\opts}.
The geometry of the final CAD model, usually stored as a B-Rep, is obtained through these operations, which makes each face of the B-Rep directly related to a type of operation. In Figure~\ref{fig:seg-group}\textcolor{red}{c}, we show some intermediate steps of CAD construction and how the faces of the corresponding B-Rep are obtained using different \opts. For example, the B-Rep of a cube that was obtained by sketching a 2D square and applying an extrusion operation, as in Figure~\ref{fig:seg-group}\textcolor{red}{a}, would result in two faces with \textit{`extrude~end'} labels and four faces with \textit{`extrude~side'} labels. The ability to automatically infer the \opt~that allowed for the creation of each face of the B-Rep constitutes a first, yet essential, step towards relating the geometry of the CAD model to its construction history. Recently introduced models~\cite{brepnet,uvnet} proposed to learn the segmentation of B-Rep faces into \opts.\\ 
Formally, let us consider a B-Rep $\mathcal{B}$ labelled with the per-face \opts~$\mathbf{T}~=~[\mathbf{t}_1;\mathbf{t}_2;\dots;\mathbf{t}_{N_f}]~\in~{\{0,1\}}^{N_f \times k_t}$, where $k_t$ is the number of possible \opts. Here, $\mathbf{T}~\in~{\{0,1\}}^{N_f \times k_t}$ is an $N_f \times k_t$ matrix with binary entries, where each row $\mathbf{t}_j \in \{0,1\}^{k_t}$ can have only one element as $1$ representing the \opt~of the face $f_j$. The task of \opt~segmentation consists of learning a mapping~$\mathbf{\Phi}$, such that, 

\setlength\abovedisplayskip{0pt}

\begin{align}
\mathbf{\Phi} : \mathbb{R}^{N_f \times d_f} \times \mathbb{R}&^{N_e \times d_e} \times \mathbb{R}^{N_c \times d_c} \rightarrow {\{0,1\}}^{N_f \times k_t} \ , \\
%\forall \: &1 \leq i \leq M, \quad
&\mathbf{\Phi}(\mathbf{F},\mathbf{E},\mathbf{C})=\mathbf{T} \ .
\end{align}
\setlength\abovedisplayskip{0pt}
It is important to highlight that the segmentation task of \opts~uses the features of faces, edges and co-edges, but assigns a unique \opt, among a fixed number of possible types, to each face of the B-Rep. Despite its usefulness for reconstructing the CAD construction history of B-Reps, the segmentation into \opts~is not sufficient as it does not describe the relationship between the faces nor the steps of the construction.   

\subsection{CAD Operation Steps} \label{sec:group-memb}

In addition to the operation types that are assigned to the faces of the B-Reps, our aim is to relate them further to the construction history. Accordingly, we propose a novel task consisting of segmenting the faces of B-Reps into CAD operation steps. For notation simplicity, they will be denoted as \textbf{\opss} in what follows. 
While the segmentation into \opts~aims at identifying the operation that was used to create each face, the purpose of the segmentation into \opss~is to group faces that were created at the same time step. An example is shown in Figure~\ref{fig:seg-group}\textcolor{red}{b}. \\
Formally, let us consider a B-Rep  $\mathcal{B}$ labelled with the per-face \opss~$\mathbf{S} \in \{0,1\}^{N_f \times k_{s}}$, where $k_s$ denotes the number of \opss~in $\mathcal{B}$. Similarly to the \opts~$\mathbf{T}$, the \opss~are represented by an $N_f \times k_s$ binary matrix $\mathbf{S}~=~[\mathbf{s}_1; \mathbf{s}_2; \dots ; \mathbf{s}_{N_f}]~\in~\{0,1\}^{N_f \times k_{s}}$. Each row of this matrix, $\mathbf{s}_j \in \{0,1\}^{k_{s}}$, can have only one element equal to $1$ denoting the \ops~for the face $f_j$. Segmenting the faces of B-Reps into \opss, would require learning a mapping~$\mathbf{\Psi}$, 

\setlength\abovedisplayskip{0pt}
\begin{align}
\mathbf{\Psi} : \mathbb{R}^{N_f \times d_f} \times \mathbb{R}&^{N_e \times d_e} \times \mathbb{R}^{N_c \times d_c} \rightarrow {\{0,1\}}^{N_f \times k_s} \ , \\
%\forall \: &1 \leq i \leq M, \quad 
&\mathbf{\Psi}(\mathbf{F},\mathbf{E},\mathbf{C})=\mathbf{S} \ .
\end{align}
\setlength\abovedisplayskip{0pt}
The proposed segmentation into \opss~is a challenging task for two main reasons: (1) unlike the \opt~segmentation where the possible types are predefined, the labels of \opss~$\textbf{S}$ are arbitrary and any combination of labels, in which faces belonging to the same step have identical labels, can be considered as correct; (2) predicting \opss~aims at grouping B-Rep faces according to the design history. Therefore, it requires learning the relationship between the different faces of the B-Rep in addition to its geometry and topology.

%----------------------------------------

%%%%%%%%%%%%%%%%%%%%%%%%%%%%%%%%%%%%%%%%%%%%%%%%%%%%%%

%----------------------------------------
\section{Proposed CADOps-Net}
\label{sec:approach}
The proposed \cadops~jointly learns the \opt~and \ops~segmentation within the same model. In practice, the mappings $\mathbf{\Psi}$ and $\mathbf{\Phi}$, introduced in Section~\ref{sect:pblmstatement}, are learnt using an end-to-end neural network. BRepNet~\cite{brepnet} is used as the backbone of our model, as it has been shown to effectively operate on B-Reps. BRepNet uses the face, edge, and co-edge features $(\mathbf{F},\mathbf{E},\mathbf{C})$ of a B-Rep $\mathcal{B}$ to learn per-face embeddings using a succession of convolutions defined through specific topological walks and Multilayer Perceptron (MLP) layers. For more details about this backbone, readers are referred to~\cite{brepnet}. In what follows, the BRepNet backbone will be denoted by $\mathbf{\Delta}~:~\mathbb{R}^{N_f \times d_f}~\times~\mathbb{R}^{N_e \times d_e}~\times~\mathbb{R}^{N_c \times d_c}~\rightarrow~\mathbb{R}^{N_f \times d_{emb}}$ and $\mathbf{f}^{\Delta}$ will be used as a notation for the embedding extracted using this backbone from a face $f$ of a B-Rep $\mathcal{B}$. The proposed network is composed of two modules that are described below.

\subsection{CAD Operation Step Segmentation} \label{section:step_module}

The CAD operation step module has two roles. Firstly, it predicts the per-face \ops~labels. Secondly, it is used to aggregate the embeddings of faces belonging to the same step and produce embeddings for each group of faces obtained in a single \ops.

\vspace{0.12cm}
\noindent\textbf{Learning CAD operation steps: }
The mapping $\mathbf{\Psi}$ introduced in Section~\ref{sec:group-memb} consists of two components, \ie, $\mathbf{\Psi}~:=~\mathbf{\sigma}~\circ~\mathbf{\Delta}$, where $\mathbf{\Delta}$ uses the features of the B-Rep $(\mathbf{F},\mathbf{E},\mathbf{C})$ and extracts per-face embeddings $\mathbf{F}^{\Delta}~=~[\mathbf{f}_1^{\Delta};~\mathbf{f}_2^{\Delta}~;~\dots~;~\mathbf{f}_{N_f}^{\Delta}]~\in~\mathbb{R}^{N_f \times d_{emb}}$. $\sigma$ is an MLP followed by softmax that maps the face embeddings $\mathbf{F}^{\Delta}$ into probabilities of predicted \opss~$\mathbf{\widehat{S}}~=~[\mathbf{\widehat{s}}_1; \mathbf{\widehat{s}}_2;~\dots~;~\mathbf{\widehat{s}}_{N_f}]~\in~{[0,1]}^{N_f \times k_{s}}$. Here, each face $f_j$ would have a vector $\mathbf{\widehat{s}}_j \in [0,1]^{k_{s}}$ specifying its membership probabilities to the $k_s$ \opss. It is important to note that the number of \opss~in a CAD model is not known in advance. We assume the maximum number of steps, $k_s$, in a B-Rep $\mathcal{B}$ to be the largest number of possible steps per model computed on the training dataset. 

As mentioned in Section~\ref{sec:group-memb}, a particular challenge for predicting the \opss~is that the ground truth labels $\mathbf{S}$ are arbitrary. Therefore, the task consists of predicting the combination of steps that matches the ground truth labels. Inspired by~\cite{spfn,point2cyl}, we use a Hungarian matching~\cite{hungarian} to find the best one-to-one correspondences between the predicted \opss~$\mathbf{\widehat{S}}$ and ground truth labels $\mathbf{S}$. Even though the Hungarian matching is not differentiable, it is only used to find the correspondences in the training phase, allowing for the computation of a \textit{Relaxed Intersection over Union} (RIoU)~\cite{param_learning} metric between pairs of predictions $\mathbf{\widehat{s}}$ and ground truth $\mathbf{s}$ as follows, 

\begin{equation}
  \mbox{RIoU}(\mathbf{s}, \mathbf{\widehat{s}}) = \frac{\mathbf{s}^{\text{T}}\mathbf{\widehat{s}}}{||\mathbf{s}||_1+||\mathbf{\widehat{s}} ||_1 -\mathbf{s}^{\text{T}}\mathbf{\widehat{s}} } \ , 
\end{equation}
where $||.||_1$ denotes the $\ell_1$ norm, and $^{\text{T}}$ the vector transpose. The RIoU metric is further used to define the following \ops~loss function, 

\begin{equation}
\mathcal{L}_{step}= \frac{1}{N_f}\displaystyle \sum_{j=1} ^{N_f} (1- \mbox{RIoU}(\mathbf{s}_j, \mathbf{\widehat{s}}_{j})) \ .
\end{equation}

\noindent For inference, the Hungarian matching is not used and the predicted \opss~are given by taking the maximum probability over each $\mathbf{\widehat{s}}$.

\vspace{0.12cm}
\noindent\textbf{CAD operation step embedding}: In addition to predicting the per-face \opss~given a B-Rep, the same module is used to extract CAD step embeddings
%$S_i^{emb} = [s_1^{emb}; s_2^{emb}; \dots ; s_{k_s}^{emb}] \in \mathbb{R}^{N_f \times d_f_{emb}}$
$\{\mathbf{s}^{\mathcal{A}}_1,\mathbf{s}^{\mathcal{A}}_2,\dots,\mathbf{s}^{\mathcal{A}}_{k_s}\}$. This is achieved by aggregating the embeddings of faces predicted to belong to the same \ops. Specifically, each \ops~$\varphi$ would have an embedding $\mathbf{s}^{\mathcal{A}}_{\varphi} \in \mathbb{R}^{d_{emb}}$, such that

\begin{equation}
    \mathbf{s}^{\mathcal{A}}_{\varphi} = \underset{j=\argmax\mathbf{\widehat{S}}_{:,	\varphi}}{\mathcal{A}}  \mathbf{f}^{\Delta}_j \ ,
\label{eq:aggregation}
\end{equation}
where $\mathbf{\widehat{S}}_{:,\varphi}$ denotes the per-face predicted \ops~labels for $\varphi$, and $\mathcal{A}$ is an aggregation function that preserves the dimension of the input embeddings such as average or maximum. Finally, each face of the B-Rep will have the corresponding \ops~embedding $\mathbf{s}^{\mathcal{A}}$ according to the predicted \ops~label. These embeddings are finally stacked in a matrix $\mathbf{S}^{\mathcal{A}} \in \mathbb{R}^{N_f \times d_{emb}}$.

%----------------------------------------
\subsection{CAD Operation Type Segmentation}
The introduced mapping $\mathbf{\Phi}$ to obtain the \opt~segmentation from an input B-Rep shares the same BRepNet backbone $\mathbf{\Delta}$ used by the module of \opt~segmentation. Moreover, it uses two other mappings, $\mathbf{\gamma}$ and $\mathbf{\rho}$, where $\mathbf{\Phi}~:=~ \mathbf{\rho} \circ \gamma \circ \mathbf{\Delta}$. The mapping $\mathbf{\gamma}~:~\mathbb{R}^{N_f \times d_{emb}}
~\times~\mathbb{R}^{N_f \times d_{emb}}~\rightarrow~\mathbb{R}^{N_f \times 2d_{emb}}$ takes as input the face embeddings $\mathbf{F}^{\Delta}$ and outputs their concatenation with the corresponding step embeddings $\mathbf{S}^{\mathcal{A}}$. These concatenated embeddings are fed to an MLP with softmax which are represented by $\mathbf{\rho}~:~\mathbb{R}^{N_f \times 2d_{emb}}
%~\times~\mathbb{R}^{N_f \times d_{f^{emb}}}
~\rightarrow~{\{0,1\}}^{N_f \times k_t}$. The final \opts~$\mathbf{\widehat{T}}$ can be obtained following, 

\begin{equation}
    \mathbf{\widehat{T}} =  \mathbf{\rho}(\mathbf{F}^{\Delta}\oplus \mathbf{S}^{\mathcal{A}}) \ ,
\label{eq:concat_CADType}
\end{equation}

\noindent where $\oplus$ is the column-wise concatenation operation.
The loss function for the \opt~segmentation is computed using the cross-entropy $\mathcal{H}$ between the predicted per-face \opts~$\mathbf{\widehat{t}}$ and the ground truth labels $\mathbf{t}$, 

\begin{equation}
  \mathcal{L}_{type} = \frac{1}{N_f}\sum_{j=1}^{N_f}{\mathcal{H}(\mathbf{t}_j, \mathbf{\widehat{t}}_j)} \ . 
\end{equation}
The total loss function is the sum of the \ops~and \opt~losses, 

\begin{equation}
    \mathcal{L}_{total} = \mathcal{L}_{step} + \mathcal{L}_{type} \ .
\end{equation}

\noindent The model jointly learns to predict the per-face \opt~and \ops~labels of a CAD model given its B-Rep, with the \opt~being conditioned on the \ops.
%%%%%%%%%%%%%%%%%%%%%%%%%%%%%%%%%%%%%%%%%%%%%%%%%%%%%%
\section{CC3D-Ops dataset}
\label{dataset}
\begin{figure}[t]
     \centering

         \includegraphics[width=0.9\linewidth]{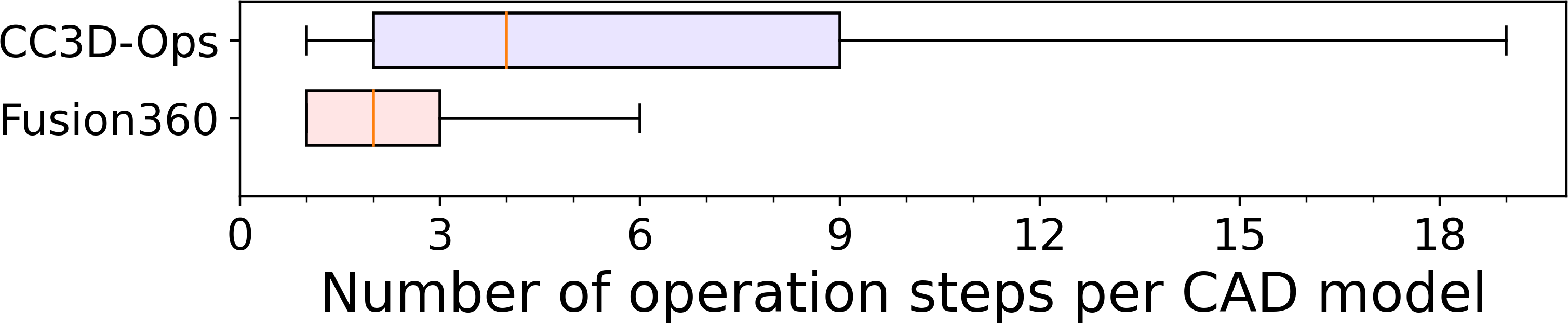}
        \caption{Boxplot of the number of \opss~per model in Fusion360~\cite{fusion360} and the proposed \ccdops~dataset.
        }
        \label{fig:dataset_stats}
\end{figure}

We introduce the \ccdops~dataset that contains $37k+$ B-Reps with the corresponding per-face \opt~and \ops~annotations. These labels were extracted using the Solidworks API~\cite{solidworks}. The B-Reps and their corresponding annotations constitute an extension of the CC3D~dataset~\cite{cc3d}.
While the Fusion360 dataset~\cite{fusion360} contains a similar number of B-Reps ($35k+$) with the corresponding \opt~labels, it does not provide \ops~labels and it includes relatively simple CAD models. The proposed \ccdops~dataset comes with more complex models that are closer to real-world industrial challenges. In Figure~\ref{fig:dataset_stats}, we illustrate the distribution of \ops~number per model as a box plot for both Fusion360 and \ccdops~datasets. It can be clearly observed that the distribution of \ccdops~is more skewed towards a higher number of \opss~than the one of Fusion360.
Specifically, {\raise.17ex\hbox{$\scriptstyle\mathtt{\sim}$}}$48\%$ of the Fusion360 models are made of only one \ops~and {\raise.17ex\hbox{$\scriptstyle\mathtt{\sim}$}}$80\%$ of them are constructed by $3$ or less \opss. On the other hand, only {\raise.17ex\hbox{$\scriptstyle\mathtt{\sim}$}}$20\%$ of the \ccdops~models are built with a single \ops~and {\raise.17ex\hbox{$\scriptstyle\mathtt{\sim}$}}$44\%$ of them with $3$ or less \opss. 
Moreover, the maximum number of \opss~per model, $k_{s}$, is $59$ for Fusion360 and $262$ for \ccdops. 
Finally, the \ccdops~dataset introduces three new \opts~to the eight present in Fusion360 which consists of, `\textit{cut revolve side}', `\textit{cut revolve end}', and `\textit{others}'. More details about the dataset can be found in the supplementary material.
\section{Experiments}
\label{sec:expt}

\subsection{Experimental Setup}
\label{sect:exp-setup}

\noindent\textbf{Input Features}: The input features of \cadops~are face, edge and co-edge features $(\mathbf{F},\mathbf{E},\mathbf{C})$ extracted from the B-Rep, $\mathcal{B}$. Following~\cite{brepnet}, the face type (\eg plane, cylinder, sphere) and area are encoded in a single vector. 3D points are further sampled on each face using the UV-grid of the B-Rep and encoded as described in~\cite{uvnet}. These two features are concatenated and used as face features. The features of the B-Rep faces are then concatenated in a \hbox{row-wise} fashion to form the matrix $\textbf{F}$. For edge features, a similar approach is taken by considering the type, convexity, closeness, length of the edge as in~\cite{brepnet}, and encoded sampled 3D points as done in~\cite{uvnet}. The result is concatenated in an edge feature matrix $\mathbf{E}$. The co-edge features, $\textbf{C}$, are simple flags to represent the direction of the corresponding edges~\cite{brepnet}.
 
\vspace{0.12cm}
\noindent\textbf{Network Architecture}: The input features are passed through a BRepNet backbone, $\mathbf{\Delta}$, with the same parameters as in~\cite{brepnet} using the wing-edge kernel. The dimension of the face embedding, $\mathbf{f}^{\Delta}$, is $d_{emb} = 64$. These embeddings are fed to an MLP followed by softmax, $\mathbf{\sigma}$, to predict the \ops. The aggregation function used to compute the step embedding, $\mathbf{S}^{\mathcal{A}}$, is the average function. Each \ops~embedding $\mathbf{s}^{\mathcal{A}}$ has the same dimension as $\mathbf{f}^{\Delta}$. The final face embedding, $\mathbf{f}^{\Delta}~\oplus~\mathbf{s}^{\mathcal{A}}$, are $128$-dimensional. Lastly, the \opt~is estimated by passing these embeddings through an MLP followed by softmax, $\mathbf{\rho}$. In our experiments, the number of layers of the employed MLPs is $1$. 

\vspace{0.12cm}
\noindent\textbf{Datasets:} \cadops~is evaluated on the Fusion360 dataset~\cite{fusion360} and the novel \ccdops~dataset described in Section~\ref{dataset}. Note that in Fusion360, the \ops~annotations were derived from the \opt~annotations as they were implicitly provided. The train, validation, and test sets for the Fusion360 dataset are the same as in~\cite{brepnet}. For the \ccdops~dataset, the splitting ratios are approximately $65\%$, $15\%$, and $20\%$ for the train, validation, and test sets. 

\vspace{0.12cm}
\noindent\textbf{Training details}: The training was conducted for $200$ epochs with a batch size of $100$ using an NVIDIA RTX $A6000$ GPU. Adam optimizer is employed with a learning rate of $0.001$ and beta parameters of $0.9$ and $0.99$.  

\vspace{0.12cm}
\noindent\textbf{Metrics}: The performance of the network is evaluated on \opt~and \ops~segmentation tasks. To evaluate the \opt~segmentation, we use the same metrics as in~\cite{brepnet}, namely, the mean accuracy (mAcc) and the mean Intersection over Union (mIoU). 
Note that we do not consider the mIoU for evaluating the \ops~as the labels represent membership sets rather than predefined classes.
Furthermore, the consistency between the \opt~and \ops~predictions is considered. For this purpose, we group the sub-\opts, such that \textit{`extrude~end'} and \textit{`extrude~side'}, into a single \textit{`extrude'} \opt. Similar grouping is done for \textit{`revolve'}, \textit{`cut~extrude'}, and \textit{`cut~revolve'}. We define an \ops~prediction as consistent if all its faces have the same \opt~prediction. To evaluate this consistency, two metrics are computed: (1) the first one, $R_{C}$, quantifies the overall consistency as the ratio of consistent predicted \opss; (2) the second one quantifies the amount of consistency of a model as $S_{C}=\sum_{i} \frac{ \max(n_{(t_1,s_i)},..., n_{(t_{k_t},s_i)} )}{n_{s_i}}$ where $n_{s_i}$ is the number of faces with \ops~label $s_i$ and $n_{(t_j,s_i)}$ the number of faces with \opt~label $t_j$ and \ops~label $s_i$. We then compute $mS_C$ as the average over all the models.

\begin{figure}[!ht]
\begin{center}
\includegraphics[trim= 5 5 5 5,clip,width=.9\linewidth]{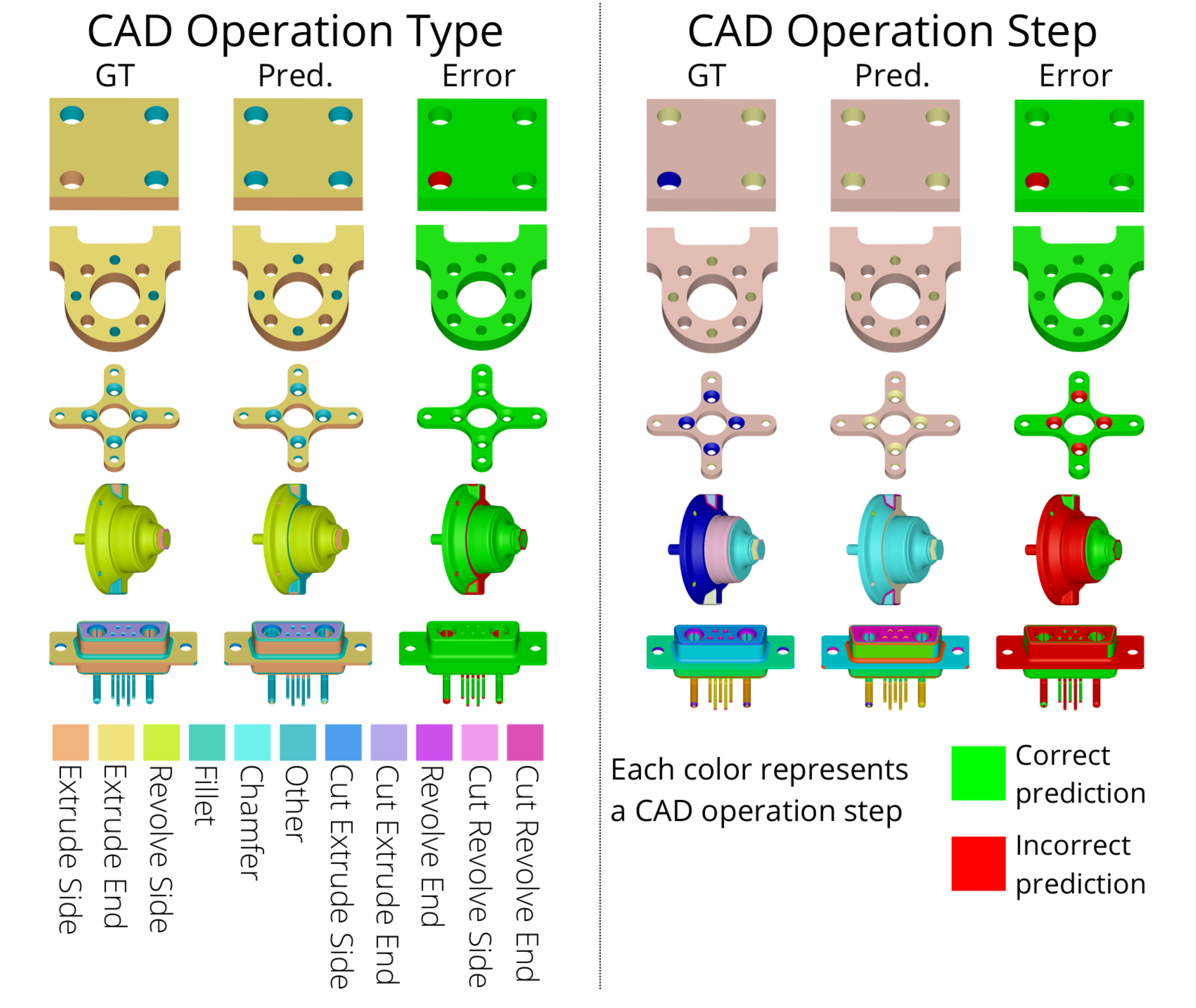}
   \caption{Sample predictions on five models from the \ccdops~dataset. (Left): The CAD operation type segmentation. (Right): The CAD operation step segmentation. For both tasks, the ground truth (GT) is shown in the left, the prediction (Pred.) in the middle, and the error (Error) in the right illustrating the correct/incorrect face predictions.}
\label{fig:prediction_examples}
\end{center}
\end{figure}
\subsection{Results and Discussions}

\noindent\textbf{Qualitative Evaluation:} In Figure~\ref{fig:prediction_examples}, we illustrate the predictions obtained by \cadops~on five models from the \ccdops~dataset. More predictions are provided in the supplementary material. Despite the complexity of some models, it can be observed that most of the \opt~predictions (left panel) were correct except for very few faces. On the other hand, the segmentation into \opss~(right panel) was more challenging for complex models (two last rows) as the segmentation into \opss~requires the model to learn the relationship between the faces of the B-Rep according to the construction history. Such aspect is more challenging to capture for complex models than the \opts~which could be hypothetically learned from the geometry and topology of the B-Reps. This hypothesis is further discussed in the quantitative evaluation.

\begin{figure}[ht]
     \centering
     
     \begin{subfigure}[b]{0.49\linewidth}
         %\centering
         \includegraphics[width=\textwidth]{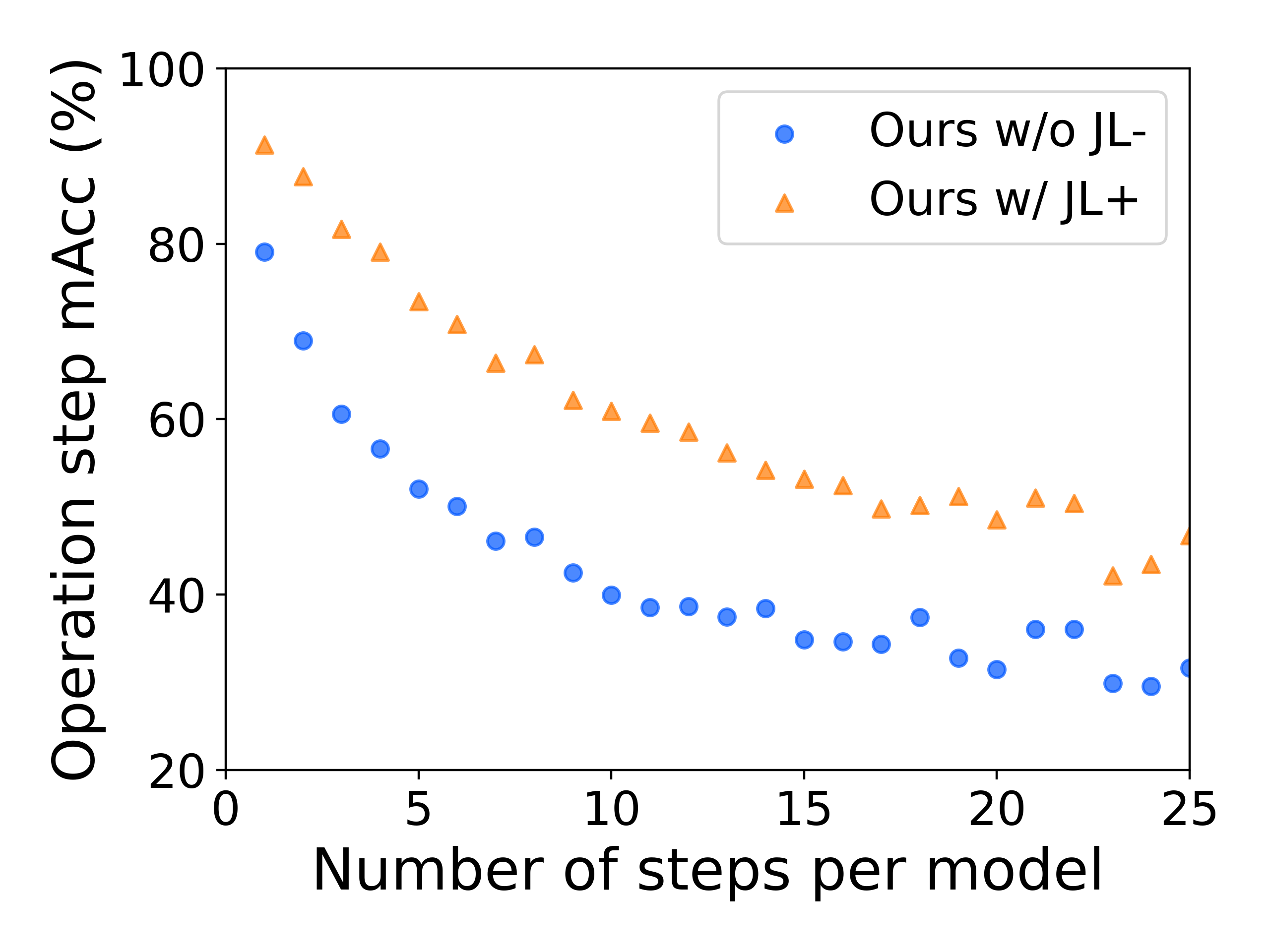}
         
         \caption{CAD operation step mAcc}
         \label{fig:group_acc}
    \end{subfigure}
    \begin{subfigure}[b]{0.49\linewidth}
         %\centering
         \includegraphics[width=\textwidth]{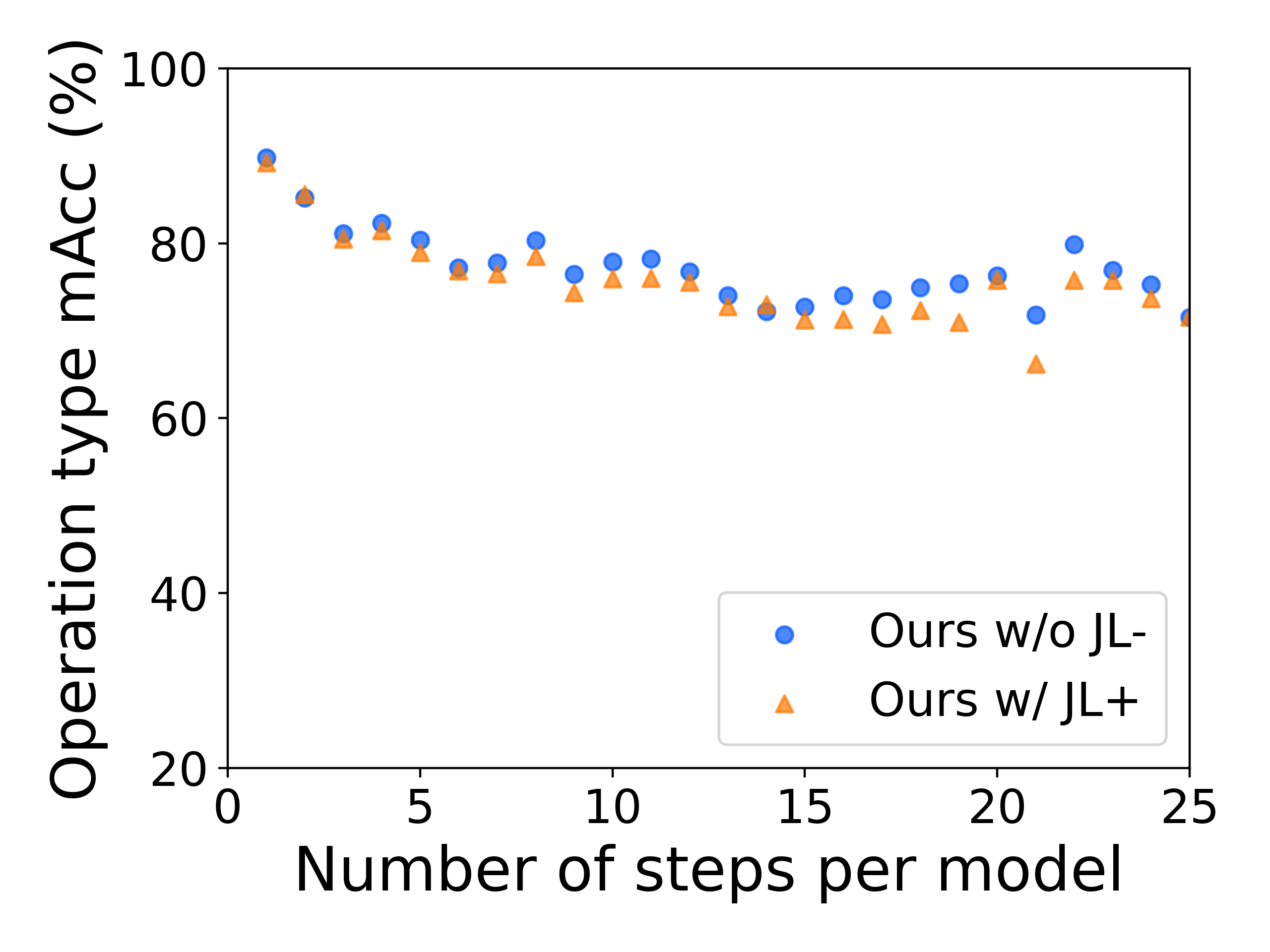}
         \caption{CAD operation type mAcc}
         \label{fig:seg_acc}
     \end{subfigure}
     \label{fig:mean_acc_cc3d}
        \caption{Mean accuracy (mAcc) of CAD operation type and step segmentation \textit{w.r.t} the number of steps per model on the \ccdops~dataset.}
\end{figure}

%\vspace{0.12cm}
\noindent\textbf{Quantitative Evaluation:} In Table~\ref{table:results}, we report the quantitative results of our approach compared to baselines. \cadops~(\textit{Ours w/ JL$^+$}) is compared to the same model without the joint learning of \opss~and \opts~(\textit{Ours w/o JL$^-$}). In the latter, the \opt~and \ops~segmentation modules are trained independently. In the following, we first analyze the results for the segmentation into \opss~(column 5 of Table~\ref{table:results}) and for the \opt~segmentation (columns 3 and 4), then we discuss the consistency between the two types of predictions (columns 6 and 7).

\begingroup
\setlength{\tabcolsep}{3pt}
\begin{table}[t!]
\centering
\begin{tabular}{cccccccc}
\hline \\[-1em]
  & \multirow{2}{*}{Model} & \multicolumn{2}{c}{\opt} & \multicolumn{1}{c}{\ops}&\multicolumn{2}{c}{\textit{\small Consistency}} \\[-1em] \\  \cline{3-4} \cline{6-7}\\[-1em]
                           &                        & mAcc         & mIoU             & mAcc         & $R_{C}$  & $mS_{C}$     \\[-1em] \\ \hline \\[-1em]
\multirow{5}{*}{\rotatebox{90}{Fusion360}} & \textit{CADNet}~\cite{cadnet}                 &   88.9               &  67.9               &       -           & -   & -               \\
                           & \textit{UV-Net}~\cite{uvnet}                 & 92.3            & 72.4           & -                & -   & -             \\
                           & \textit{BRepNet}~\cite{brepnet}                & 94.3            & 81.4           & -            & -   & -         \\
                           \cline{2-7}
                           & \textit{Ours w/o JL$^-$  }           & 95.5            & 83.2           & 80.2            & 87.1 & 97.4  \\
                           & \textit{Ours w/ JL$^+$}                   & \textbf{95.9}   & \textbf{84.2}  & \textbf{82.5}   & \textbf{93.3} & \textbf{98.7}   
                           \\[-1em]\\ \hline \\[-1em]
                           
\multirow{4}{*}{\rotatebox{90}{\ccdops}}  & \textit{CADNet}~\cite{cadnet}                 &  57.5                &    26.9             &    -              & -     & -             \\
                           & \textit{BRepNet}~\cite{brepnet}                & 71.4            & 35.9           & -            & -  & -          \\
                           \cline{2-7}
                           & \textit{Ours w/o JL$^-$}             & \textbf{76.0}   & 43.0           & 48.4            & 40.7 & 82.7           \\
                           & \textit{Ours w/ JL$^+$}                   & 75.0            & \textbf{44.3}  & \textbf{62.7}   & \textbf{82.4} & \textbf{96.7}  \\[-1em]
                           \\ \hline
\end{tabular}
\caption{Results of the segmentation into CAD operation types and steps on the Fusion360 and \ccdops~datasets. All results are expressed as percentages. \textit{Ours w/o JL$^-$} denotes our method without joint learning. \textit{Ours w/ JL$^+$} refers to the proposed \cadops~with joint learning.}
\label{table:results}
\end{table}
\endgroup
%\vspace{-0.5cm}

 As previously mentioned, predicting \opss~is a much more challenging task than \opts~especially for models with a large number of \opss.
While the joint learning leads to small improvements on the \ops~mAcc metric on the Fusion360 dataset, significant improvements can be observed on the \ccdops~dataset results with an increase of {\raise.17ex\hbox{$\scriptstyle\mathtt{\sim}$}}$14\%$. This difference of results can be explained by the higher complexity of \ccdops~models compared to those of Fusion360. Figure~\ref{fig:group_acc} shows the mAcc of \ops~segmentation related to the number of \opss~per model on the \ccdops~dataset. It can be observed that for models with less than $25$ \opss, representing over $96\%$ of the \ccdops~dataset, \cadops~scores consistently and significantly better than without joint learning. These observations demonstrate the importance of the joint learning for \ops~segmentation. However, in both cases there is a major decrease in the \ops~segmentation mAcc as the number of steps per model increases. This is expected since the task becomes increasingly challenging as the number of \opss~becomes larger. Note that we did not compare our results to state-of-the-art (BRepNet~\cite{brepnet}, UV-Net~\cite{uvnet}, and CADNet\cite{cadnet}) on the task of \ops~segmentation as their methods are not designed to predict arbitrary face labels.

In order to evaluate the \opt~segmentation of \cadops, the results are compared to state-of-the-art results. On the Fusion360 dataset, we recorded slight improvements over~\cite{brepnet},~\cite{uvnet}, and~\cite{cadnet} in terms of mAcc. More significant improvements \textit{w.r.t}~\cite{uvnet} and~\cite{cadnet} were obtained in terms of mIoU (more than $12\%$ and $16\%$, respectively). On the \ccdops~dataset, our results clearly outperformed those of~\cite{brepnet} and~\cite{cadnet} on the two metrics. Furthermore, we compare \cadops~to the scenario where the joint learning is omitted. One interesting observation is that there is no significant difference between the two scenarios. The same observation holds in Figure~\ref{fig:seg_acc} where we show the mAcc of \opt~segmentation related to the number of \opss~per model. In contrast to the \ops~segmentation, one can notice that the number of \opss~has a slight impact on the \opt~mAcc. In other words, the \opt~segmentation does not become more challenging when complex models with large number of construction steps are involved. Intuitively, it can be hypothesized that the \opt~segmentation is more related to the geometry and topology of the B-Rep rather than its construction history. 

The results on the consistency scores ($R_{C}$ and $mS_{C}$) highlight the relevance of the joint learning approach. Despite relatively similar \opt~and \ops~mAcc scores on the Fusion360 dataset for \textit{Ours w/ JL$^+$} and \textit{Ours w/o JL$^-$}, the joint learning approach produces more consistent results with an increase of {\raise.17ex\hbox{$\scriptstyle\mathtt{\sim}$}}$6\%$ in $R_C$ score. Similarly on the \ccdops~dataset, the predictions from \cadops~are significantly more consistent with an increase of {\raise.17ex\hbox{$\scriptstyle\mathtt{\sim}$}}$41\%$ in $R_C$ score and $14\%$ in $mS_C$ score. Therefore, the joint learning model is able to extract face features that contain consistent information for both the \opt~and \ops~segmentation labels. The consistency property is essential for the process of reverse engineering.

%%%%%%%%%%%%%%%%%%%%%%%%%%%%%%%%%%%%%%%%%%%%%%%%%%%%%%
%----------------------------------------
\subsection{Ablation Study}
\begin{table}[ht]
\centering
\begin{tabular}{ccccc}
\hline
                          &                             & \multicolumn{2}{c}{\opt} & \multicolumn{1}{c}{\ops} \\ \cline{3-4} 
                          & Agg. type                       & mAcc            & mIoU           & mAcc                      \\ \hline
\multirow{5}{*}{\rotatebox{90}{\ccdops}} & \textit{No agg.}     & 73.0            & 40.2           & 61.5                      \\
                          & \textit{Soft labels} & 73.4            & 40.0           & 59.7                     \\
                          & \textit{Sum}                & 70.4            & 34.4           & 62.6                      \\
                          & \textit{Max}                & 74.3            & 42.0            & 62.2             \\
                          & \textit{Avg}            & \textbf{75.0}   & \textbf{44.3}  & \textbf{62.7}              \\ \hline
\end{tabular}
\caption{Ablation study on the aggregation function used in the joint learning of \cadops. All results are expressed as percentages.}
\label{table:ablation}
\end{table}
In order to provide a deeper insight into the joint learning approach, we conduct an ablation study on the aggregation function $\mathbf{\mathcal{A}}$ of the face embeddings. Experiments are conducted with the following five scenarios: (1) the output face embeddings, $\mathbf{f}^{\Delta}$, from the BRepNet backbone are directly used to predict both the \opt~and \ops~without any aggregation (\textit{No agg.}). (2) Another scenario concatenates the BRepNet face embeddings with the predicted soft labels of the \ops~(\textit{Soft labels}) again without any aggregation. (3) The last three scenarios focus on the type of aggregation function used to obtain the \ops~embeddings, $\mathbf{S}^{\mathcal{A}}$, namely the maximum (\textit{Max}), the average (\textit{Avg}), and the sum of the embeddings combined with a softmax normalization (\textit{Sum}). Table~\ref{table:ablation} shows the ablation results for both \opt~and \ops~segmentation tasks on the \ccdops~dataset. The results show that aggregating the face embeddings using an \textit{Avg} pooling leads to slightly better overall performance.

%----------------------------------------

\subsection{CAD Sketch Recovery}
\begin{figure}[t]
    \centering
	\includegraphics[width=0.82\linewidth]{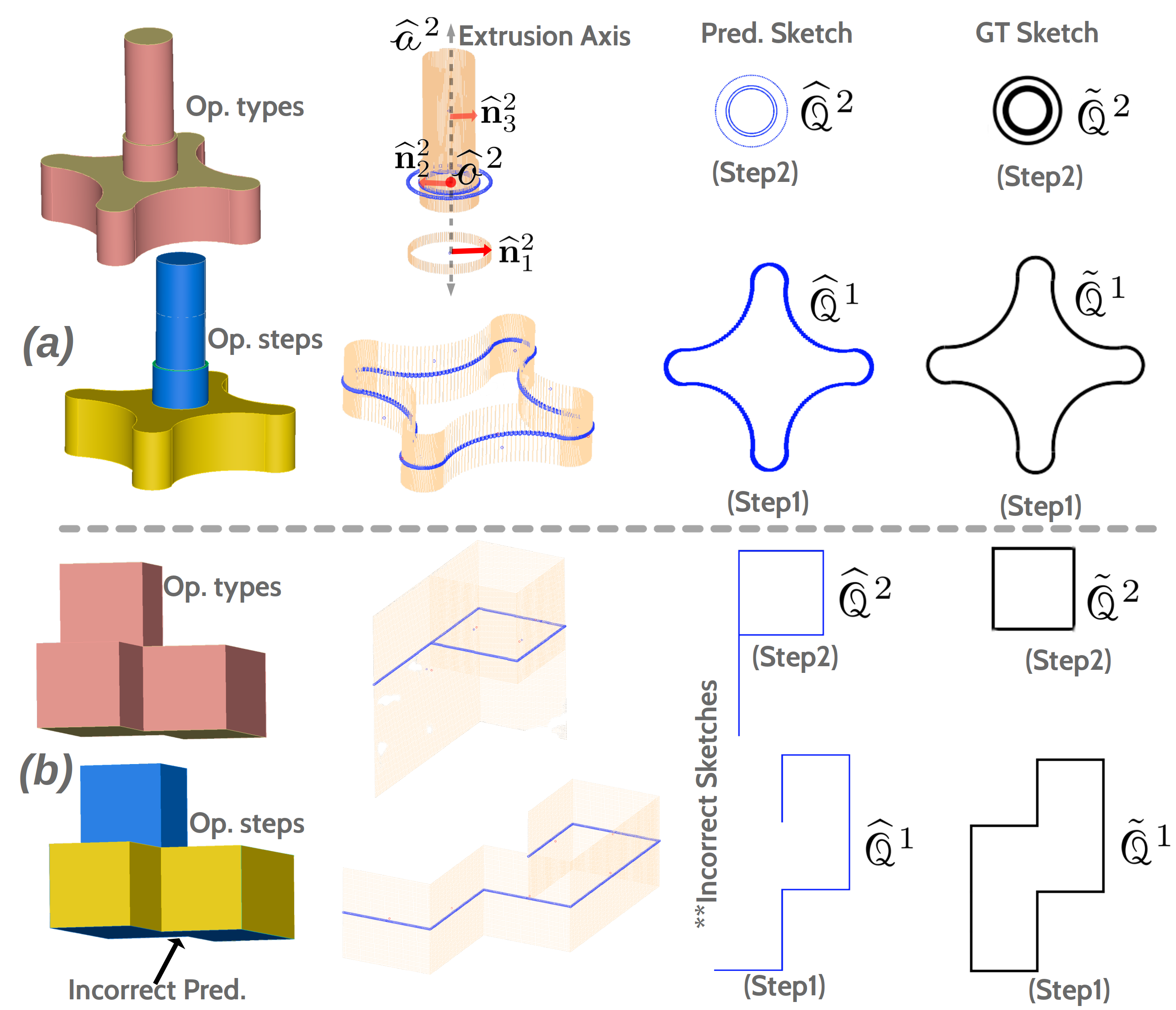}
    \caption{Sketch recovery from predicted CAD operation types (\opts) and steps (\opss). \ops~1 and 2 are colored in yellow and blue, respectively. Figure \ref{fig:prediction_examples} defines the color codes used for different \opts.
    }
    \label{fig:SketchFinal}
\end{figure}
Figure~\ref{fig:SketchFinal} illustrates preliminary results on how \cadops~predictions can be used to retrieve the CAD sketches.
A sketch $\mathbf{\mathcalboondox{Q}}$ of a B-Rep $\mathcal{B}$ can be defined as a set of simple geometrical entities (\eg straight lines, arcs). We consider a small subset of $20$ models made of extrusions from the Fusion360 dataset. In the following, we describe the process for recovering the sketch corresponding to \ops~$2$ using the \cadops~predictions shown in Figure~\ref{fig:SketchFinal}\textcolor{red}{a}. We first identify the faces for which the \opt~was predicted as `\textit{extrude side}'. Second, we cluster these faces according to their predicted \ops. Third, we store the face-normals  ($\mathbf{\widehat{n}}_{1}^{2},\hdots, \mathbf{\widehat{n}}_{m}^2$) and sample UV-grid points on the faces. This allows to derive a common \textit{axis of extrusion} $\widehat{\mathbf{\mathcalboondox{a}}}$ and a \textit{projection center} $\widehat{\mathbf{\mathcalboondox{o}}}$. Finally, the predicted sketch $\mathbf{\widehat{\mathcalboondox{Q}}}^{2}$ is obtained by projecting the sampled points along $\widehat{\mathbf{\mathcalboondox{a}}}$ (more details are in the supplementary material). Figure~\ref{fig:SketchFinal}\textcolor{red}{a} and \ref{fig:SketchFinal}\textcolor{red}{b} show qualitative results of successful and failed sketch recoveries from correctly and incorrectly predicted \opts. These preliminary results on sketch recovery illustrate the relevance of \ops~prediction in the context of 3D reverse engineering.

\subsection{Limitations}
\begin{figure}[t]
\begin{center}
\includegraphics[width=0.9\linewidth]{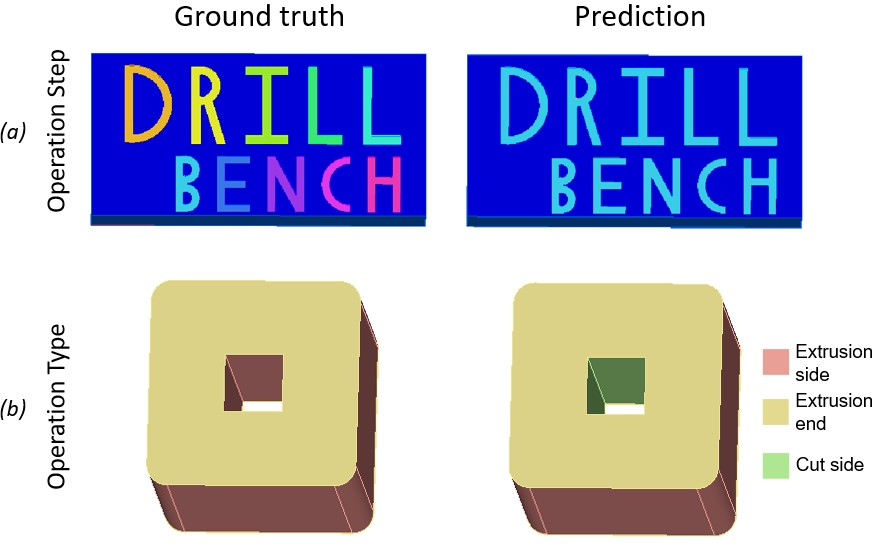}
   \caption{Failure cases of \cadops~for \ops~segmentation in (a) and \opt~segmentation in (b).}
\label{fig:group_wrong_pred}
\end{center}
\end{figure}

In CAD modeling, designers may opt for different design solutions. Consequently, the segmentation into \opt~and \ops~is not necessarily unique.
An example for which the \ops~prediction is valid despite not matching the ground truth can be found in Figure~\ref{fig:group_wrong_pred}\textcolor{red}{a}. The letters were predicted as part of the same \ops, which could be a valid design approach. However, these letters were extruded with separate \opss~in the ground truth. In Figure~\ref{fig:group_wrong_pred}\textcolor{red}{b}, an example with valid predictions of \opts~not matching the ground truth is depicted. Here, the hole in the center of the shape was predicted as a `\textit{cut}' type operation, while being an `\textit{extrude}' in the ground truth. In general, CAD designers follow good practices so that the final model reflects the design intent~\cite{design_intent}. However, different designers might have their own set of good practices, making it difficult for a learning-based model to capture all the different design intents.

%%%%%%%%%%%%%%%%%%%%%%%%%%%%%%%%%%%%%%%%%%%%%%%%%%%%%%
\section{Conclusion}
\label{sec:conclusion}
In this work, \cadops, a neural network that jointly learns the CAD operation type and step segmentation of B-Rep faces is presented. The joint learning strategy leads to significantly better results for the challenging task of CAD operation step segmentation, while achieving state-of-the-art results on the CAD operation type segmentation task. Moreover, we showed the potential of combining these two segmentations for recovering further information of the construction history such as sketches. Finally, the \ccdops~dataset is introduced with its operation type and step annotations. We believe that this dataset will help in advancing research on CAD modeling thanks to the complexity of the CAD models. As future work, an investigation of the ordering of the construction steps while maintaining various types of CAD operations would allow for the recovery of a more complete construction history.

\vspace{0.3cm}
\noindent \textbf{Acknowledgement:} The present project is supported by the National Research Fund, Luxembourg under the BRIDGES2021/IS/16849599/FREE-3D and IF/17052459/CASCADES projects, and by Artec 3D.
%%%%%%%%%%%%%%%%%%%%%%%%%%%%%%%%%%%%%%%%%%%%%%%%%%%%%%

{\small
\bibliographystyle{ieee_fullname}
\bibliography{egbib}
}

%%%%%%%%%%%%%%%%%%%%%%%%%%%%%%%%%%%%%%%%%%%%%%%%%%%%%%
%%%%%%%%%%%%%%%%%%%%%%%%%%%%%%%%%%%%%%%%%%%%%%%%%%%%%%
%%%%%%%%%%%%%%%%%%%%%%%%%%%%%%%%%%%%%%%%%%%%%%%%%%%%%%

\newpage
\setcounter{section}{0}
\setcounter{figure}{0}
\setcounter{table}{0}
\setcounter{page}{1}
\begin{center}
  \textbf{\Large Supplementary Material to the Paper: CADOps-Net: Jointly Learning CAD Operation Types and Steps from Boundary-Representations}
  \vspace{8pt}
  \end{center}

\begin{figure*}[h!]
     \centering
         \includegraphics[width=\linewidth]{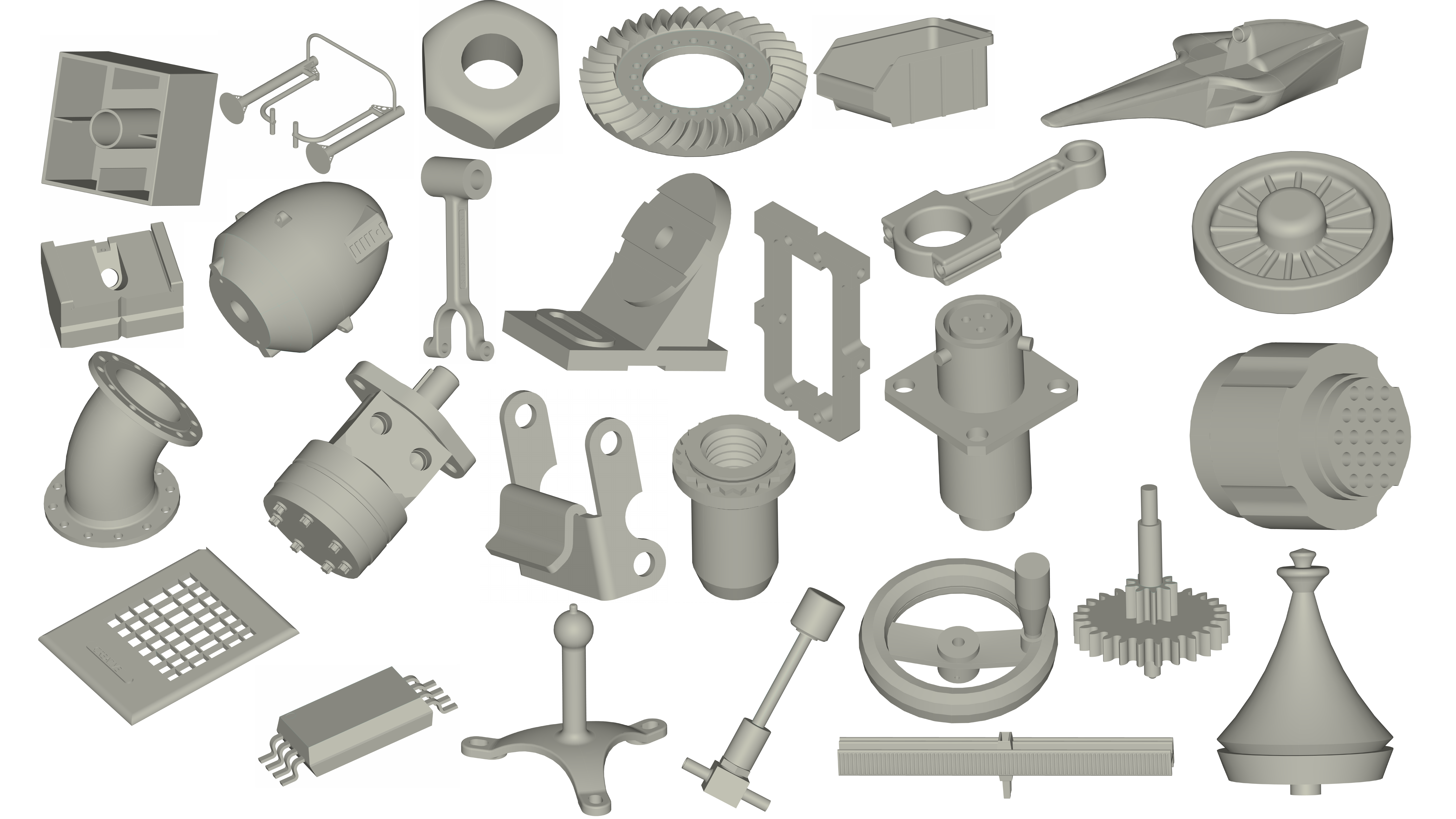}
         \caption{Sample CAD models from the \ccdops~dataset.}
         \label{fig:cc3d_sample}
\end{figure*}
%%%%%%%%% BODY TEXT
\section{CC3D-Ops Dataset}

In this section, we provide more details on the proposed \ccdops~dataset. 
First, the method used to extract the \ops~and \opt~will be briefly discussed.
Then, statistics demonstrating the complexity of the models and showing the distribution of the labels will be presented. 

\subsection{CC3D-Ops Label Extraction}
The proposed \ccdops~dataset contains B-Reps with per-face \opt~and \ops~annotations. An important aspect of segmenting faces into different construction steps of modeling operations is that these labels come from the real construction history of each CAD model in the dataset. In our case, this information is obtained from the native SolidWorks~\cite{solidworks} Part File (.sldprt) format of a CAD model.  
A set of tools were developed based on the Solidworks API~\cite{solidworks} to traverse a CAD model's construction history and to assign each face generated by respective modeling operation its \opt~and \ops~labels in B-Rep.

\subsection{Statistics}

\begin{figure}[h!]
     \centering
         \includegraphics[width=\linewidth]{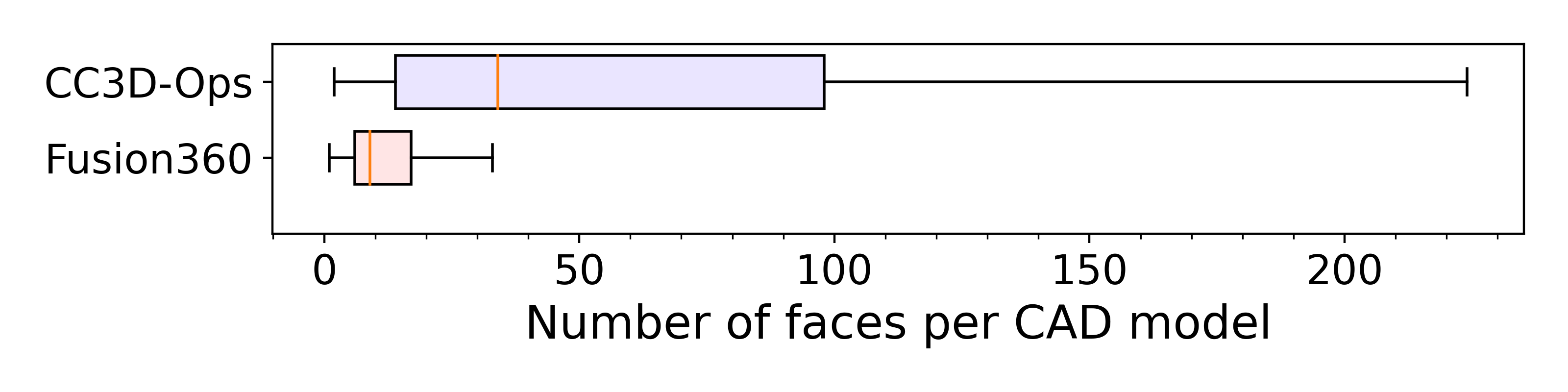}
         \caption{Box plot showing the distribution of models for the \ccdops~and Fusion360~\cite{fusion360} datasets with respect to the number of faces per model.}
         \label{fig:face_box}
\end{figure}

\begin{figure}[h!]
     \centering
         \includegraphics[width=\linewidth]{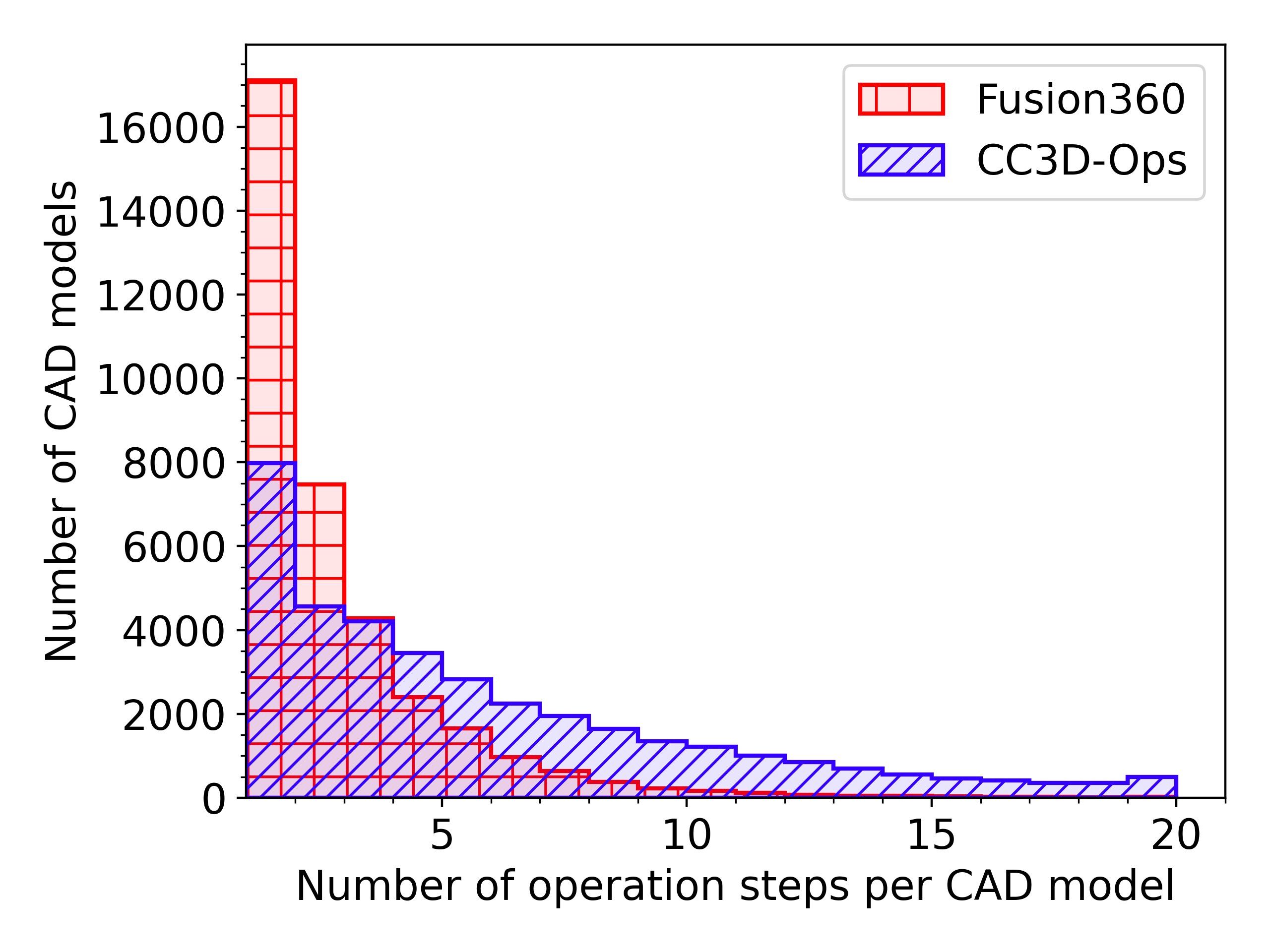}
         \caption{Histogram showing the distribution of models for the \ccdops~and Fusion360~\cite{fusion360} datasets with respect to the number of \opss~per model. Models with a number of \opss~between $0$ and $20$ represent over $96\%$ of \ccdops.}
         \label{fig:step_hist}
\end{figure}

\noindent \textbf{Model complexity:} From the sample \ccdops~CAD models (in B-Rep format) displayed in Figure~\ref{fig:cc3d_sample}, it can be noted that \ccdops~offers a wide variety of models both in terms of complexity and category. Figure~\ref{fig:face_box} shows the distribution of the number of faces per model for the \ccdops~and Fusion360~\cite{fusion360} datasets as box plots. This figure shows that the models in \ccdops~generally have more faces than in Fusion360. While $90\%$ of the models of the Fusion360 dataset have $30$ faces or less, such models represent only $50\%$ of \ccdops. This difference between the two datasets is further demonstrated in Figure~\ref{fig:step_hist}, where it can be observed that the models in \ccdops~tend to be made of more CAD operation steps than for Fusion360.

\begin{figure}
     \centering
         \includegraphics[width=\linewidth]{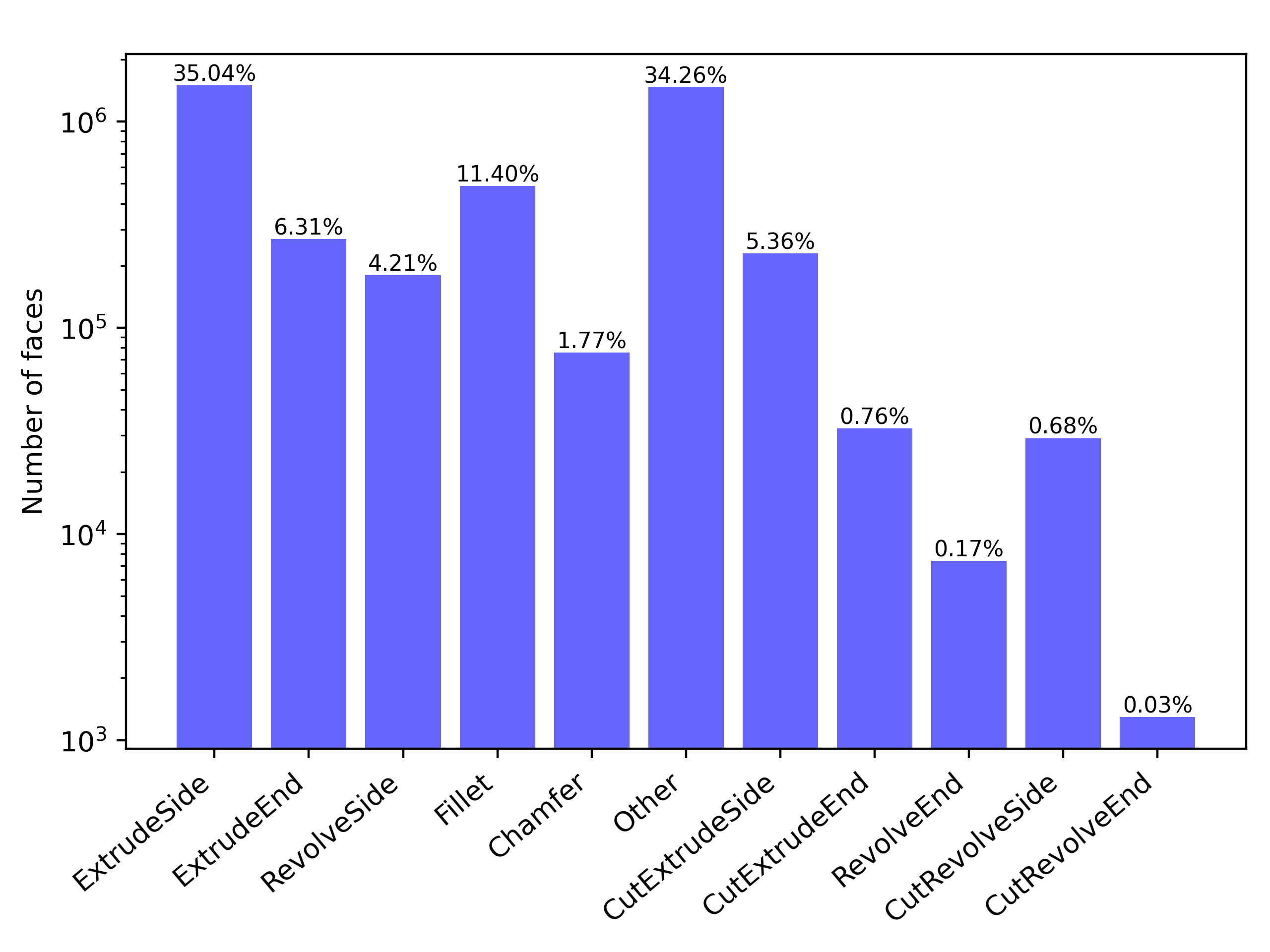}
         \caption{Bar graph of the number of faces for each \opt~ label over the \ccdops~dataset. The numbers above each bar represents the percentage of the number of faces with the corresponding type. Note: a \textit{log} scale is used for the vertical axis.}
         \label{fig:cc3d_type_barchart}
\end{figure}

\vspace{0.12cm}
\noindent \textbf{CAD Operation Type Labels:} The \opt~face labels indicate the type of CAD operation used during the design process. While the most common CAD operation types (such as \textit{extrusion}, \textit{fillet} ...) are shared among most CAD software applications, some are software specific. The \ccdops~dataset contains $11$ different \opt~labels: \textit{extrude side}, \textit{extrude end}, \textit{revolve side}, \textit{revolve end}, \textit{cut extrude side}, \textit{cut extrude end}, \textit{cut revolve side}, \textit{cut revolve end}, \textit{fillet}, \textit{chamfer} and \textit{other}. The \textit{other} \opt~represents less common types such as \textit{helix}, \textit{sweep}, \textit{dome}, \etc. The bar chart in Figure~\ref{fig:cc3d_type_barchart} displays the number of faces for each \opt~label. The two least common \opt~labels are \textit{revolve end} and \textit{cut revolve end} and the two most common operation types are \textit{extrude side} and \textit{other}. For a comparison with the Fusion360 dataset, we refer the reader to~\cite{brepnet} where a similar bar chart can be found.

%%%%%%%%%%%%%%%%%%%%%%%%%%%%%%%%%%%%%%%%%%%
\section{Further Experimental Analysis}

In this section, we first analyze quantitative results on the per class IoUs for the \opt~segmentation task. Then, some visual examples of the \cadops~predictions are presented.

\subsection{CAD Operation Type IoUs}

As in~\cite{brepnet}, an analysis of each \opt~IoU is presented. Table~\ref{table:Fusion_IoUs} and~\ref{table:CC3D_IoUs} show the IoU results of each \opt~label for the Fusion360 and \cadops~datasets respectively. The results are shown for both \cadops~without joint learning (\textit{Ours~w/o~JL$^-$}) and with joint learning (\textit{Ours~w/~JL$^+$}). As noted in Section~\textcolor{red}{6.2} of the main paper, the joint learning strategy does not have a significant impact on the \opt~predictions. This is particularly the case on the Fusion360 dataset as shown in Table~\ref{table:Fusion_IoUs}. \textit{Ours~w/~JL$^+$} achieves slightly higher IoU for each class. On the other hand, the same trend cannot be found in the results obtained from the \ccdops~dataset. For $6$ out of the $11$ \opt~classes, the difference between the IoUs obtained with joint learning and without is relatively small (less than $2\%$). For the \opt~classes \textit{cut revolve side} and \textit{chamfer}, \textit{Ours~w/o~JL$^-$} scores higher than the \textit{Ours~w/~JL$^+$} by $3.9\%$ and $5.5\%$ respectively. However, the joint learning method achieves higher results on $3$ out of the $4$ least common classes, namely \textit{cut revolve end}, \textit{revolve end} and \textit{cut extrude end}. In particular, for the \textit{revolve end} \opt~that represents $0.17\%$ of the dataset, the joint learning strategy results in an IoU that is about $17\%$ higher than without joint learning. This demonstrates that even if the joint learning strategy achieves a comparable mIoU as without joint learning, \textit{Ours~w/~JL$^+$} is able to learn more meaningful features for the underrepresented \opts.

\begin{table}[h!]
\centering
\begin{tabular}{lcc}
\hline
 Fusion360            & \multicolumn{2}{c}{Per class IoU}            \\ \cline{2-3} 
             & \textit{Ours w/o JL$^-$} & \textit{Ours w/ JL$^+$} \\ \hline
\textit{Extrude side} & 94.0                   & \textbf{94.6}       \\
\textit{Extrude end}  & 91.7                   & \textbf{92.4}       \\
\textit{Cut side}     & 82.1                   & \textbf{83.9}       \\
\textit{Cut end}      & 75.2                   & \textbf{77.1}       \\
\textit{Revolve side} & 85.1                   & \textbf{86.5}       \\
\textit{Revolve end}  & 48.7                   & \textbf{48.9}       \\
\textit{Chamfer}      & 91.2                   & \textbf{92.1}       \\
\textit{Fillet}       & 97.6                   & \textbf{97.8}       \\ \hline
\end{tabular}
\caption{\opt~per class IoU for the Fusion360 dataset. All results are expressed as percentages.}
\label{table:Fusion_IoUs}
\end{table}

\begin{table}[h!]
\centering
\begin{tabular}{lcc}
\hline
    \ccdops                      & \multicolumn{2}{c}{Per class IoU}            \\ \cline{2-3} 
                          & \textit{Ours w/o JL$^-$} & \textit{Ours w/ JL$^+$} \\ \hline
\textit{Extrude side}     & \textbf{65.4}          & 64.9                \\
\textit{Extrude end}      & 59.7                   & \textbf{60.2}       \\
\textit{Cut extrude side} & 17.8                   & \textbf{18.1}       \\
\textit{Cut extrude end}  & 10.0                   & \textbf{15.4}       \\
\textit{Cut revolve side} & \textbf{22.2}          & 18.3                \\
\textit{Cut revolve end}  & 1.1                    & \textbf{4.6}        \\
\textit{Revolve side}     & \textbf{60.3}          & 59.8                \\
\textit{Revolve end}      & 23.8                   & \textbf{41.2}       \\
\textit{Chamfer}          & \textbf{69.6}          & 64.4                \\
\textit{Fillet}           & \textbf{84.1}          & 83.1                \\
\textit{Other}            & \textbf{58.5}          & 57.2                \\ \hline
\end{tabular}
\caption{\opt~per class IoU for the \ccdops~dataset. All results are expressed as percentages.}
\label{table:CC3D_IoUs}
\end{table}

%%%%%%%%%%%%%%%%%%%%%%%%%%%%%%%%%%%%%%%%%%%

\subsection{CAD Operation Types Qualitative Results}

\begin{figure*}[h!]
     \centering
         \includegraphics[trim= 5 5 5 5,clip, width=0.85\textwidth]{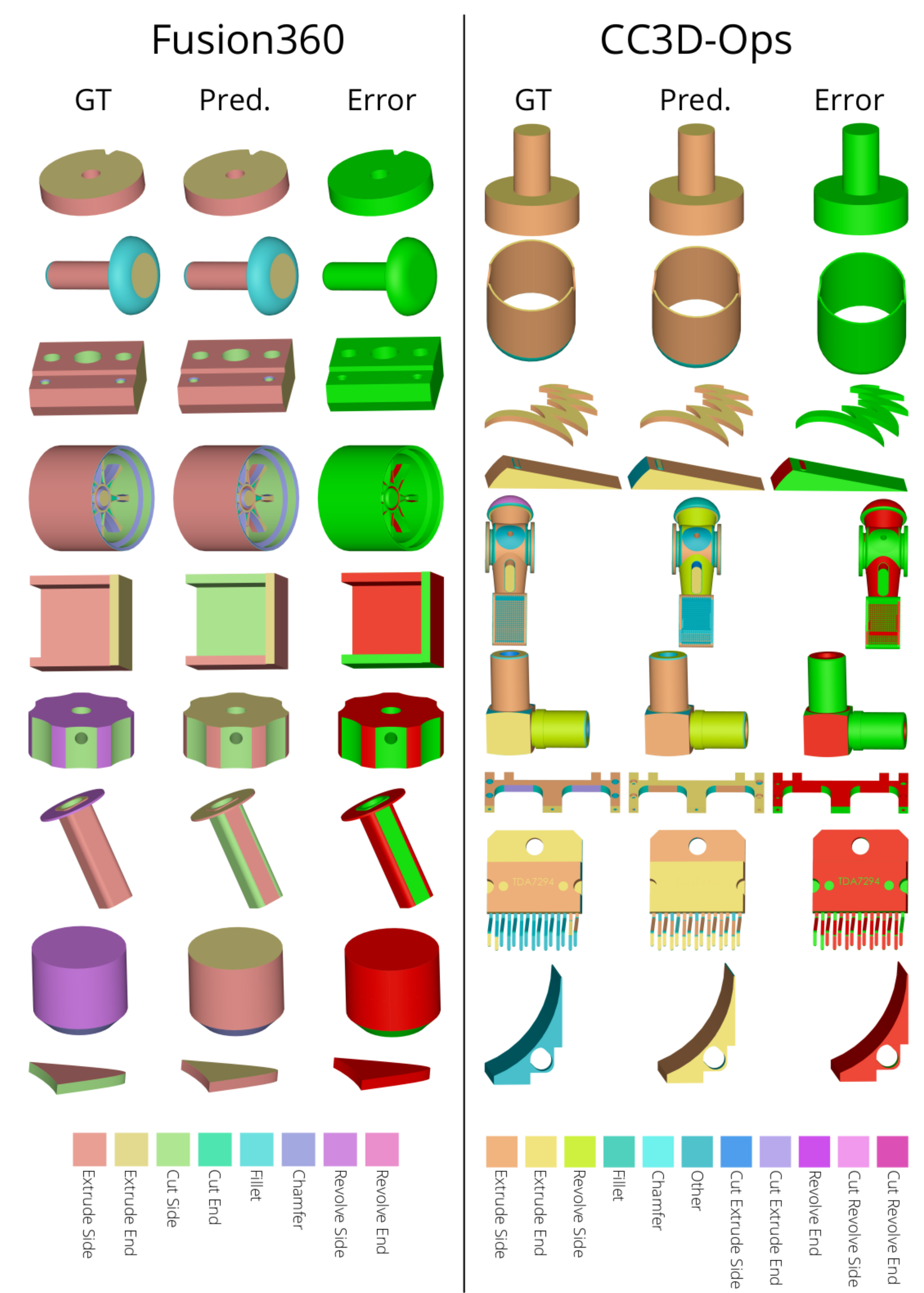}
         \caption{CAD operation step qualitative results on the Fusion360 (left) and \ccdops~(right) datasets. For the \cadops~ground truth (GT) and predictions (Pred.). Correctly segmented faces are shown in green and incorrect in red in the Error columns.}
         \label{fig:type_qualitative}
\end{figure*}

Figure~\ref{fig:type_qualitative} shows sample results from \cadops~for the \opt~segmentation task on both the Fusion360 and \ccdops~datasets. As discussed in Section \textcolor{red}{6.2} of the main paper, the \opt~accuracy is not correlated to the complexity of the models. In particular, this can be observed from the results presented on the Fusion360 dataset in which \cadops~sometimes fails to predict the correct \opts~for some simple models. Despite not matching the ground truth, some predictions are still valid within the context of CAD modelling, as outlined in Section~\textcolor{red}{6.5}.

\subsection{CAD Operation Step Qualitative Results}
\begin{figure*}[h!]
     \centering
         \includegraphics[trim= 5 5 5 5,clip,width=0.95\textwidth]{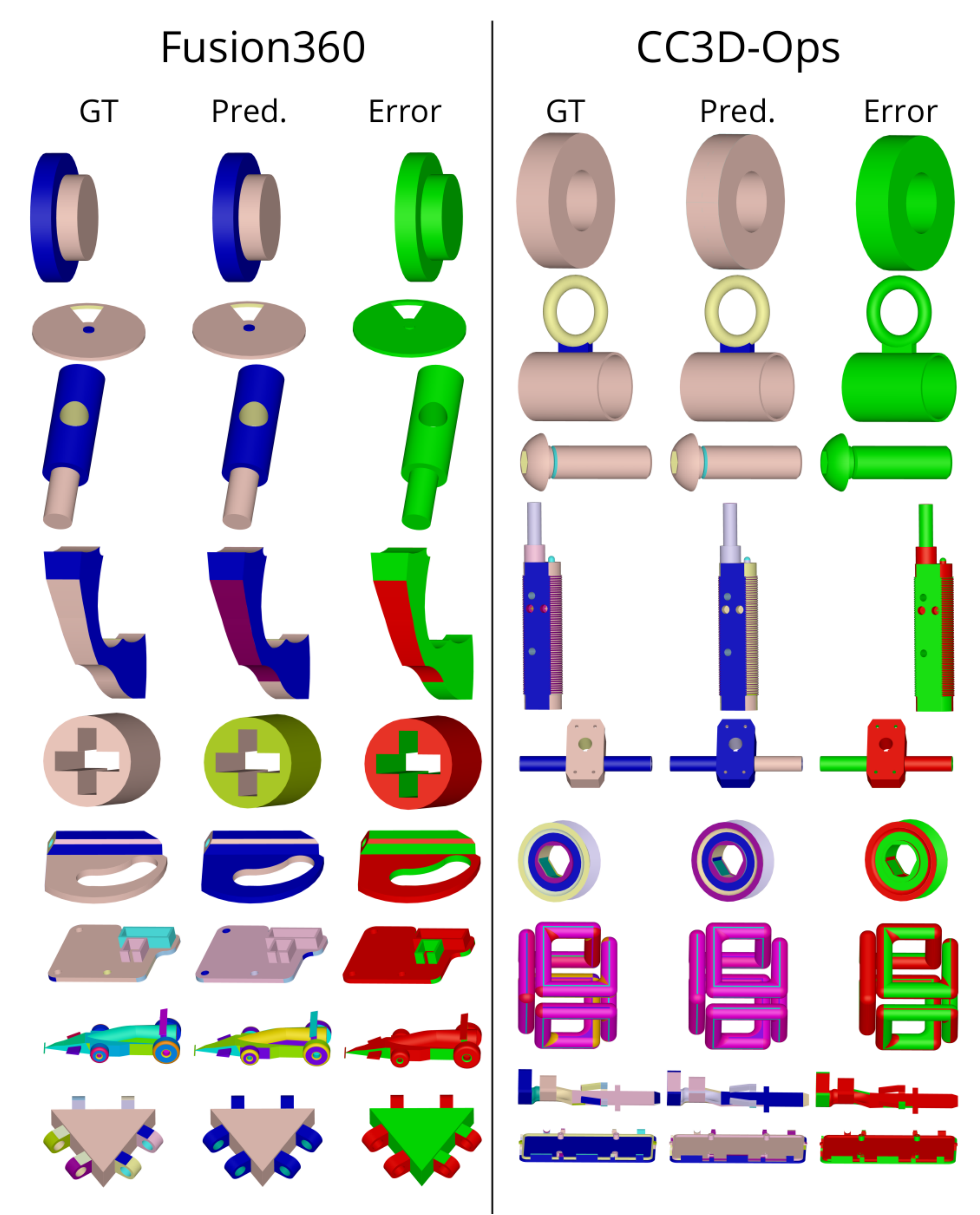}
         \caption{CAD operation step qualitative results on the Fusion360 (left) and \ccdops~(right) datasets.For the \cadops~ground truth (GT) and predictions (Pred.), each color represents a CAD operation step. Correctly segmented faces are shown in green and incorrect in red in the Error columns.}
         \label{fig:step_qualitative}
\end{figure*}

Figure~\ref{fig:step_qualitative} displays qualitative results for the \ops~segmentation task on the \ccdops~and Fusion360 datasets. These qualitative results illustrate that while \cadops~is able to make accurate predictions for models with a small number of \opss, the \ops~accuracy decreases as the number of \opss~increases, as explained in Section~\textcolor{red}{6.2}.

%%%%%%%%%%%%%%%%%%%%%%%%%%%%%%%%%%%%%%%%%%%
\section{Sketch Recovery Process and Examples}\label{sec:Insights_SketchRecovery}
\RestyleAlgo{ruled}
\SetKwInput{KwInput}{Input}
\SetKwInput{KwOutput}{Output}
\SetKwComment{Comment}{/* }{ */}
\begin{algorithm}[!ht]
%\begin{algorithmic}[1]
\caption{Sketch Recovery Algorithm}\label{alg:SkectchRecovery}
\KwInput{$\mathcal{B}, \widehat{\mathbf{S}}\in~{[0,1]}^{N_f \times k_{s}}, 
\widehat{\mathbf{T}} \in~{[0,1]}^{N_f \times k_{t}}$}
\KwOutput{a set of sketches $  \mathbf{\mathcalboondox{Q}}$}
\ForEach{\text{id} $\mathbf{s} \in \{ \operatorname*{argmax} \widehat{\mathbf{S}}_{j:}, \,\, \forall  j \in [1,\hdots,N_f] \} $}{%
    \ForEach{\text{id} $\mathbf{t} \in \{ \operatorname*{argmax} \widehat{\mathbf{T}}_{j:}, \,\, \forall j \in [1,\hdots,N_f]\}$}{%
    $\mathbf{\mathcalboondox{F}} \gets \{..,\mathcalboondox{f}_{\mathbf{s}}, ...\}$ \Comment*[l]{Group face ids ($\mathbf{s,t}$) into $\mathcalboondox{f}_{\mathbf{s}}$ of \opt~\lq extrude side\rq at different \opss}
    }
}
\ForEach{$\mathcalboondox{f}_{\mathbf{s}} \in \mathbf{\mathcalboondox{F}}$}{%
compute normals $\{\widehat{\mathbf{n}}_i^{\mathbf{s}}, \, \forall i\in \mathcalboondox{f}_{\mathbf{s}}\}$\;
$\widehat{\mathcalboondox{o}}_{\mathbf{s}} \gets $  centroid of merged faces $\in \mathcalboondox{f}_{\mathbf{s}}$\; 
$\widehat{\mathcalboondox{a}}_{\mathbf{s}} \gets $ optimal extrusion axis $ \sum_{\substack{i,j \in \mathcalboondox{f}_{\mathbf{s}} \\ i\neq j}} \frac{\widehat{\mathbf{n}}_i^{\mathbf{s}} \times \widehat{\mathbf{n}}_j^{\mathbf{s}}}{\lvert\widehat{\mathbf{n}}_i^{\mathbf{s}} \times \widehat{\mathbf{n}}_j^{\mathbf{s}}\rvert}$\;
$\widehat{\mathcalboondox{Q}}^{\mathbf{s}} \gets $ project merged faces $\in \mathcalboondox{f}_{\mathbf{s}}$ on a plane with center $\widehat{\mathcalboondox{o}}_{\mathbf{s}}$ and axis $\widehat{\mathcalboondox{a}}_{\mathbf{s}}$  
}
%\end{algorithmic}

\end{algorithm}
\begin{figure*}[ht!]
     \centering
         \includegraphics[width=0.99\textwidth]{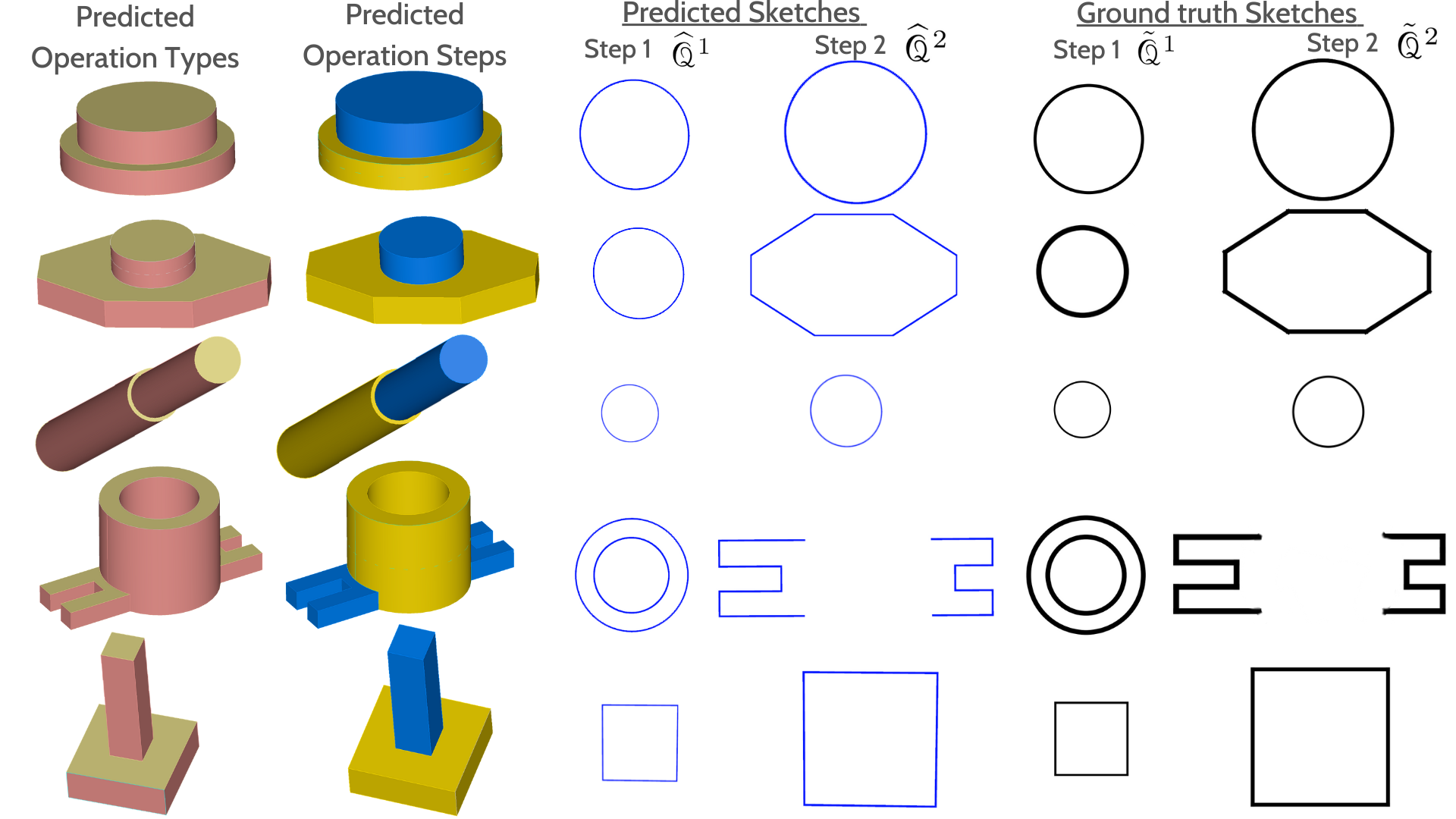}
         \caption{Qualitative results on sketch recovery from correctly predicted~\opts~ and~\opss~by CADOps-Net. The models shown above include exactly two operation steps.}
         \label{fig:SketchesPred_Suppl}
\end{figure*}
\begin{figure*}[ht!]
     \centering
         \includegraphics[width=0.99\textwidth]{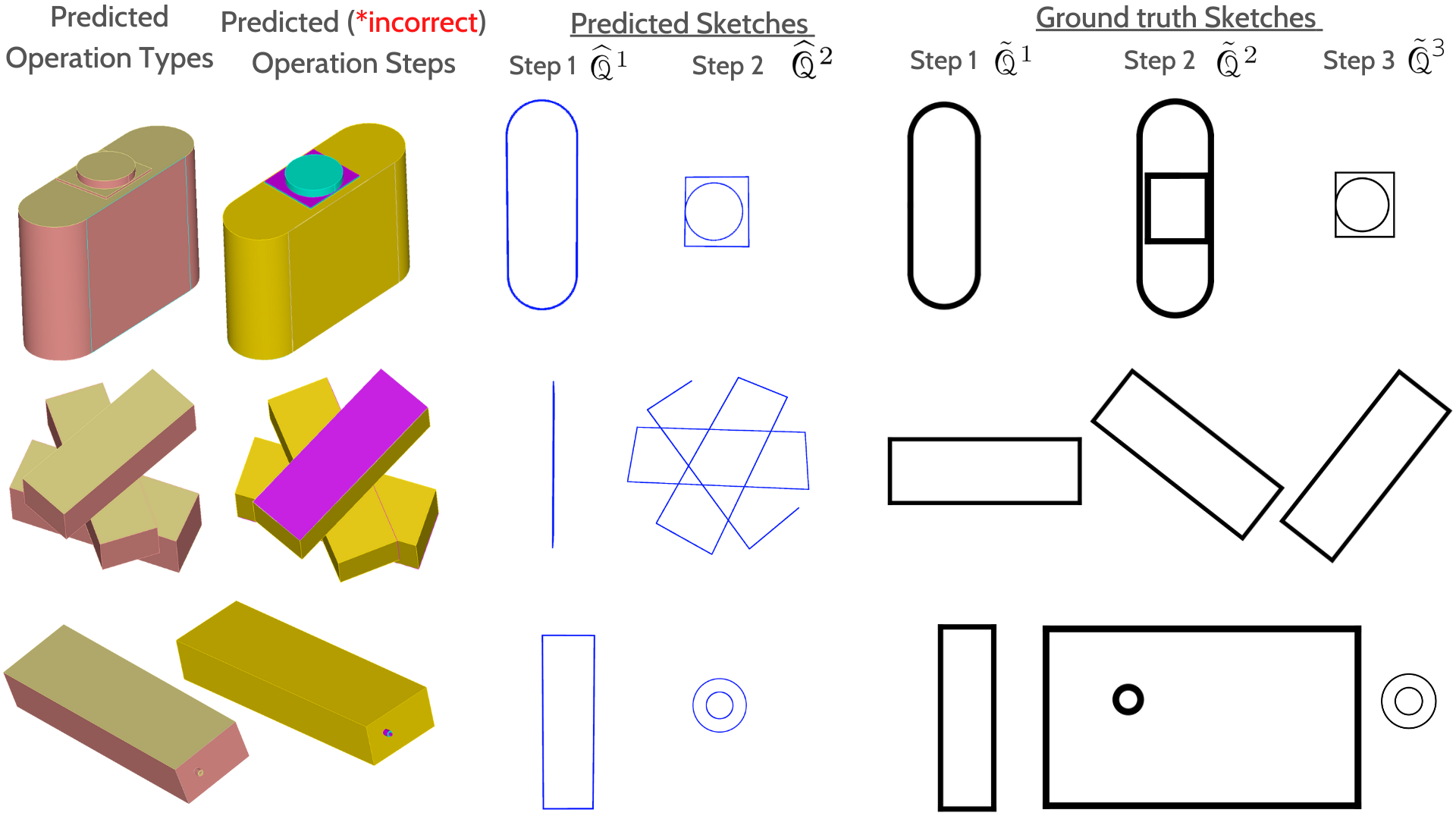}
         \caption{Qualitative results on sketch recovery from incorrectly predicted \opts~and \opss~by CADOps-Net. The models shown above include exactly three operation steps.}
         \label{fig:SketchesPredIncorrect_Suppl}
\end{figure*}

In CAD modeling, sketches are considered as starting points to build a kernel structure (in our case B-Rep) of the desired solid model.
Nevertheless, the sketches %involved in such modeling 
are only part of the forward design process, and tracing them back from the final B-Rep is not straightforward. In this section, we provide more details about the process of sketch recovery introduced in Section~\textcolor{red}{6.4} and present more qualitative results.

\newtheorem{prop}{Proposition}
\begin{prop}
\label{prop:No-Coplanar_Profile}
If we assume that not all faces of a B-Rep are either orthogonal or parallel to each other, then the base sketch-profile of the merged faces may not be co-planar, and therefore extrusion or revolution axis may not be orthogonal to the normals of co-faces.
\end{prop}

Algorithm~\ref{alg:SkectchRecovery} describes the process for retrieving sketches at different \opss~using \cadops~predictions ($\mathbf{\widehat{S}}$ and $\mathbf{\widehat{T}}$), assuming the proposition \ref{prop:No-Coplanar_Profile} holds over a B-Rep when extrusion-only operation is involved. In particular, \cadops~predictions of \opss~are used to group the faces of the B-Rep that were created by a single sketch. Among these faces, the ones created by \textit{extrude side} are identified by the  \opt~predictions (lines 1 to 5 of Algorithm~\ref{alg:SkectchRecovery}). These faces are denoted $\mathcalboondox{f}_{\mathbf{s}}$ and are considered to compute the extrusion axis and the centroid, then the projection plane as described in lines 6 to 11 of Algorithm~\ref{alg:SkectchRecovery}.

Figure~\ref{fig:SketchesPred_Suppl} and 
\ref{fig:SketchesPredIncorrect_Suppl} illustrate the qualitative results of sketches recovered using \cadops~predictions and Algorithm~\ref{alg:SkectchRecovery} on randomly selected samples made by extrusions from Fusion360~\cite{fusion360} dataset. We dissect the sketch results based on correctly and incorrectly predicted \opss~in the two figures. 
We can observe that the sketch recovery is successful when the predictions of \opss~are correct (Figure~\ref{fig:SketchesPred_Suppl}). In Figure~\ref{fig:SketchesPredIncorrect_Suppl}, the B-Reps were segmented into two \opss, while the ground truth annotations indicated that they were designed through three \opss. Such incorrect \ops~prediction impacted the sketch recovery and resulted in erroneous sketches. 

\end{document}